%% file: sidecar_SISC.tex
\documentclass[hidelinks,onefignum,onetabnum]{siamart250211}

\input{ex_shared}

\begin{document}

\maketitle
\begin{abstract}
 Solving partial differential equations (PDEs) with neural networks (NNs) has shown great potential in various scientific and engineering fields.
However, most existing NN solvers mainly focus on satisfying the given PDE formulas in the strong or weak sense, without explicitly considering some intrinsic physical properties, such as mass and momentum conservation, or energy dissipation. This limitation may result in nonphysical or unstable numerical solutions, particularly in long-term simulations. To address this issue, we propose ``Sidecar'', a novel framework that enhances the physical consistency of existing NN solvers for solving parabolic PDEs. Inspired by the time-dependent spectral renormalization approach, our Sidecar framework introduces a small network as a copilot, guiding the primary function-learning NN solver to respect the structure-preserving properties.
Our framework is highly flexible, allowing the preservation of various physical quantities for different PDEs to be incorporated into a wide range of NN solvers.
Experimental results on some benchmark problems demonstrate significant improvements brought by the proposed framework to both accuracy and structure preservation of existing NN solvers.

\end{abstract}



\section{Introduction}

Partial differential equations (PDEs) are fundamental tools for modeling physical systems, including fluid dynamics, electromagnetism, and quantum mechanics.
Since most PDEs do not have analytical solutions, numerical methods are developed to obtain approximate solutions with good accuracy and efficiency.
In addition to satisfying the target  PDEs in the strong or weak sense, numerical solutions should also consistently respect the intrinsic physical properties of the systems, such as mass conservation,  energy dissipation, and maximum bounds.
Therefore, it is essential and highly desirable for numerical solvers to incorporate structure-preservation into the solution process to ensure the stability and physical fidelity of the resulting approximate solutions.
Such principles have been extensively studied in traditional numerical methods, where structure-preserving properties are embedded into the scheme design, leading to robust and reliable results \cite{Christiansen_Munthe-Kaas_Owren_2011, du2021maximum,leveque1992numerical,QX2024structure, SHARMA2020113067}.

Recently, neural network (NN)-based methods have become powerful and flexible numerical tools in science and engineering.
With advancements in hardware and algorithms, NNs can effectively learn intricate patterns and representations of PDE solutions,
enabling the development of various NN-based solvers \cite{Zhoutao2023Failure, Zhaojia2021Solving, pang2019fpinns, raissi2019physics, wang2024respecting, wang2021understanding, yu2018deepritz, zang2020WAN}.
These function-learning solvers train NNs to approximate PDE solutions directly from the PDE formulas, bypassing the need for high-resolution training data.
This makes NN solvers particularly advantageous for problems with high dimensionality or complex geometries, where conventional numerical approaches often encounter significant difficulties.

However, most existing NN solvers mainly focus on exploiting the given PDE formulas (e.g., minimizing the PDE residuals), without explicitly accounting for the intrinsic physical properties of the system. This limitation may result in nonphysical solutions, reducing the stability capability of NN solvers.
Recent works have attempted to incorporate the structure-preserving information into NN solvers \cite{geng2024deep, Hernandez2021Structure, huang2024mass, kutuk2024energy}.
Ideally, the structure-preserving properties should facilitate the learning process of NN solvers rather than hinder them.
However, a common issue of existing methods is the undesired trade-off between accuracy and physical fidelity, ending up with a performance drop.
Existing structure-preserving methods for NN solvers can mainly be divided into two categories: \textbf{a)} \emph{hard constraints}: to manually post-process or project the network's outputs to enforce the physical structure \cite{geng2024deep, Hernandez2021Structure}, and \textbf{b)} \emph{soft regularization}: to introduce additional regularization terms into the loss function \cite{huang2024mass, kutuk2024energy}.
Specifically, hard constraints methods often suffer from distribution shifts between training and testing data, which can hurt the generalization ability.
Meanwhile, soft regularization methods often encounter difficulties due to the need for numerical integration of preserved quantities, which often involve spatial integrals. Since NN solvers are trained using gradient-based optimization, the loss function must be differentiable. This requirement makes it challenging to use standard numerical integration schemes, as differentiable integration is not always practical, especially in the high-order schemes.

In this paper, we propose a novel framework, ``Sidecar", designed to enhance the preservation of physical properties of existing NN solvers for parabolic PDEs.
The key idea is to introduce a lightweight network as a copilot, guiding the primary function-learning NN solver to respect the structure-preserving properties.
Both NNs are trained simultaneously: the primary network approximates the PDE solution, while the copilot network maintains the needed physical properties via an additional structure loss term.
This flexible and plug-and-play design enables Sidecar to cooperate with various NN solvers and adapt to different types of PDEs with diverse structure-preserving properties.
The copilot network design is inspired by the Time-Dependent Spectral Renormalization (TDSR) method \cite{cole2017tdsr, hou2024energy}, which is a traditional structure-preserving approach for solving parabolic PDEs.

However, implementing TDSR in NN solvers faces notable challenges, particularly in incorporating structure knowledge and collaborating between the two networks.
These challenges are addressed through tailored loss function implementation and training procedures.

We integrate Sidecar with existing NN solvers such as physics-informed neural network (PINN) \cite{raissi2019physics} and its extensions \cite{kutuk2024energy, wang2024respecting}, and then test their performance through some benchmark PDEs, including conservative systems (nonlinear Schrödinger equation) and dissipative systems (Allen--Cahn equation).
The resulting solutions show significant enhancement in both accuracy and physical properties.
It demonstrates that Sidecar effectively incorporates structure information into NN solvers without sacrificing performance.
A further discussion provides insights into the advantages of the key components of Sidecar, including the network architecture and loss functions.
We believe that this work showcases the potential of integrating traditional numerical methods with NN-based approaches to develop more accurate and efficient solvers.

The rest of this paper is organized as follows:
\Cref{sec:preliminaries} introduces some preliminaries of structure-preserving properties and NN solvers for parabolic PDEs.
\Cref{sec:methodology} presents the detailed design of the proposed framework, Sidecar, and its application to solving parabolic PDEs with structure-preservation.
\Cref{sec:experiments} evaluates the numerical performance of Sidecar on some benchmark problems.
\Cref{sec:discussions} presents further discussions and ablation studies.
Finally, \Cref{sec:conclusion} outlines the conclusion and future research directions.

\section{Preliminaries}
\label{sec:preliminaries}

In this section, we introduce the general idea of structure-preservation for parabolic PDEs, the TDSR method, and the PINN method with some extensions, which serve as the foundation of the proposed Sidecar framework.

\subsection{Structure-preserving properties of parabolic PDEs}

For a parabolic PDE system, the structure-preserving properties are often the intrinsic physical laws that the solutions satisfy.
Let us consider a general parabolic PDE problem in the following form:
\begin{equation} \label{eq:pde}
	\begin{dcases}
		u_t = \mathcal{A} [u], & \quad (\mathbf{x}, t) \in \Omega \times (0,T], \\
		u(\mathbf{x}, 0) = u_0(\mathbf{x}), & \quad \mathbf{x} \in \Omega, \\
		u(\mathbf{x}, t) = g(\mathbf{x}, t), & \quad (\mathbf{x}, t) \in \partial \Omega \times (0,T],
	\end{dcases}
\end{equation}
where $\Omega \in \mathbb{R}^d$ is the spatial domain, $T$ is the final time,  $\mathcal{A}$ is a given differential operator,
the solution $u(\mathbf{x}, t) \in \mathbb{K}$ ($\mathbb{K} = \mathbb{R}$ or $\mathbb{C}$),
and $u_0(\mathbf{x})$ and $g(\mathbf{x}, t)$ are the given initial condition and boundary condition, respectively.
Many above systems possess some intrinsic physical properties that can be characterized by the temporal evolution of certain physical quantities:
\begin{equation} \label{eq:preserved}
	\begin{dcases}
		\frac{\dif}{\dif t} \mathcal{Q}[u] = \mathcal{S}[u], \\
		\mathcal{Q}[u](0) = C_0,
	\end{dcases}
\end{equation}
where $\mathcal{Q}[u](t), \mathcal{S}[u](t) \in \mathbb{K}$ are the concerned
physical quantity and its evolution speed function with respect to the solution $u$ at time $t$, respectively.
The initial value $C_0= \mathcal{Q} \circ \iota[u_0]$ is determined by the initial condition $u_0(\mathbf{x})$, where $\iota$ is the natural embedding operator $\iota: (\Omega \to \mathbb{K}) \to (\Omega \times [0, T] \to \mathbb{K})$ at $t=0$, and $\circ$ denotes the composition of operators.

The goal is to find a numerical solution $\bar u(\mathbf{x}, t)$ that satisfies the PDE \eqref{eq:pde} as well as the structure-preserving properties \eqref{eq:preserved}, ensuring that the solutions are stable, accurate, and physically meaningful.
For clarity, we use $u$ to denote the exact solution and $\bar u$ for a general numerical solution.

\begin{example}[Conservative system] \label{ex:NLS}
	The nonlinear Schrödinger (NLS) equation is a complex-valued PDE system (\ie, $u \in \mathbb{C}$) with the form
\begin{equation} \label{eq:nls}
	\begin{dcases}
		- i u_t = \frac{1}{2} \Delta u + |u|^2 u, & (\mathbf{x}, t) \in \Omega \times [0, T], \\
		u(\mathbf{x}, 0) = u_0(\mathbf{x}), & \mathbf{x} \in \Omega, \\
		u(\mathbf{x}, t) = g(\mathbf{x}, t),  & (\mathbf{x}, t) \in \partial \Omega \times [0, T],
	\end{dcases}	
\end{equation}
where $|\cdot|$ denotes the modulus of a complex number.
The NLS equation has several conserved quantities, including the mass and momentum of the wave function.
Define the total mass of the wave function as
$$ \mathcal{Q}_1[u](t) := \int_\Omega |u(\mathbf{x},t)|^2 \, \dif \mathbf{x},$$
then the \emph{mass conservation law} describes that the total probability density of the wave function remains constant over time, which is given by
\begin{equation} \label{eq:nls_mass}
	\begin{dcases}
		\frac{\dif}{\dif t} \mathcal{Q}_1[u] = \mathcal{S}_1[u] := 0, \\
		\mathcal{Q}_1[u](0) = C_0^{(1)},
	\end{dcases}
	\end{equation}
	where
	$ C_1^{(1)} =  \mathcal{Q}_1 \circ \iota [u_0] = \int_\Omega |u_0(\mathbf{x})|^2 \dif \mathbf{x}.$
Additionally, we can define the total momentum of the wave function as
\begin{equation*}
	\mathcal{Q}_2[u](t) := \im \left[ \int_\Omega \big( \nabla u(\mathbf{x},t) \cdot u^*(\mathbf{x},t) \big) \dif \mathbf{x} \right] = \int_\Omega \Big( \! \re[u]  \im[\nabla u] - \im [u]  \re[\nabla u] \Big) \dif \mathbf{x},
\end{equation*}
where $\re[v]$ and $\im[v]$ denote the real and imaginary parts of the complex function $v(x, t)$, respectively, while $u^* = \re[u] - i \, \im[u]$ is the complex conjugate of $u$. The corresponding \emph{momentum conservation law} gives
\begin{equation} \label{eq:nls_momentum}
	\begin{dcases}
		\frac{\dif}{\dif t} \mathcal{Q}_2[u] = \mathcal{S}_w[u] :=0, \\
	\mathcal{Q}_2[u](0) = C_0^{(2)},
	\end{dcases}
	\end{equation}
where  $C_0^{(2)} =  \mathcal{Q}_2 \circ \iota [u_0]  = \int_\Omega \Big( \! \re [u_0] \im [\nabla u_0] - \im [u_0]  \re [\nabla u_0] \Big) \dif \mathbf{x}.$
Ideally, the numerical solution $\bar u(x, t)$ should satisfy both the NLS equation \eqref{eq:nls} and the conservation laws given in Eqs.~\eqref{eq:nls_mass}-\eqref{eq:nls_momentum}.

\end{example}

\begin{example}[Dissipative system] \label{ex:allen_cahn}
The Allen--Cahn (AC) equation is a typical nonlinear phase-field model for the phase transition phenomena \cite{AllenCahn}, which is given as
\begin{equation} \label{eq:allen_cahn}
	\begin{dcases}
		u_t = \varepsilon^2 \Delta u + f[u], & (\mathbf{x}, t) \in \Omega \times [0, T], \\
		u(\mathbf{x}, 0) = u_0(\mathbf{x}), & \mathbf{x} \in \Omega, \\
	\end{dcases}
\end{equation}
where $\varepsilon$ reflects the width of the transition regions, and $f[u]$ is a reaction term.
The problem is subject to suitable boundary conditions, such as Neumann or periodic boundary conditions on $\partial \Omega$.
As a gradient flow model, the AC equation satisfies the \emph{energy dissipation law}. Specifically, the energy functional is defined as
\begin{equation} \label{eq:allen_cahn_energy}
	\mathcal{E}_{AC}[u](t) := \int_\Omega \left( \frac{\varepsilon^2}{2} | \nabla u(\mathbf{x}, t)|^2 + F[u](\mathbf{x}, t) \right) \dif \mathbf{x},
\end{equation}
where $F[u]$
is the potential functional with  $-F' = f$. Therefore, the solution to \eqref{eq:allen_cahn} should decrease the energy  defined in \eqref{eq:allen_cahn_energy} over time as
\begin{equation} \label{eq:allen_cahn_dissipation}
	\begin{dcases}
		\frac{\dif} {\dif t} \mathcal{E}_{AC}[u] = \mathcal{S}_{AC}[u] ,\\
		\mathcal{E}_{AC}[u](0) = C_0,
	\end{dcases}
\end{equation}
where
$$\mathcal{S}_{AC}[u] := - \int_\Omega u_{t}^2 (\mathbf{x}, t) \dif \mathbf{x} \leq 0$$
and $C_0= \mathcal{E}_{AC} \circ \iota [u_0]=\int_\Omega \left( \frac{\varepsilon^2}{2} | \nabla u_0|^2 + F[u_0]\right) \dif \mathbf{x}.$
\end{example}

\subsection{Time-dependent spectral renormalization approach}

TDSR method \cite{cole2017tdsr, hou2024energy} is a structure-preserving technique in traditional numerical schemes for solving parabolic PDEs, which introduces a time-dependent factor to incorporate the structure equation \eqref{eq:preserved} into the original PDE problem \eqref{eq:pde}.
TDSR mainly focuses on the scenarios where the physical quantities $\mathcal{Q}[u]$ and its evolution speed $\mathcal{S}[u]$ in \eqref{eq:preserved} are global, for instance, involving integrals over the spatial domain $\Omega$.
Therefore, we can assume $\mathcal{Q}[u]$ and $\mathcal{S}[u]$ both have the integral form as
\begin{equation*}
	\begin{aligned}
		\mathcal{Q}[u](t) = \int_\Omega \mathcal{K}_{\mathcal{Q}}[u](\mathbf{x}, t) \dif \mathbf{x}, \quad
		\mathcal{S}[u](t) = \int_\Omega \mathcal{K}_{\mathcal{S}}[u](\mathbf{x}, t) \dif \mathbf{x},
	\end{aligned}
\end{equation*}
where $\mathcal{K}_{\mathcal{Q}}, \mathcal{K}_{\mathcal{S}}: (\Omega \times [0, T] \to \mathbb{K}) \to (\Omega \times [0, T] \to \mathbb{K})$ are known operators served as the integral kernels of $\mathcal{Q}$ and $\mathcal{S}$, respectively.
The structure equation \eqref{eq:preserved} is then transformed into the integral form as
\begin{equation} \label{eq:preserved_integrated}
	\begin{dcases}
		\frac{\dif}{\dif t} \int_\Omega \mathcal{K}_{\mathcal{Q}}[u](\mathbf{x}, t) \dif \mathbf{x} = \int_\Omega \mathcal{K}_{\mathcal{S}}[u](\mathbf{x}, t) \dif \mathbf{x}, \\
		\int_\Omega \mathcal{K}_{\mathcal{Q}}[u](\mathbf{x}, 0) \dif \mathbf{x} = C_0,
	\end{dcases}
\end{equation}
where $C_0 = \int_\Omega \mathcal{K}_{\mathcal{Q}} \circ \iota [u_0](\mathbf{x}) \dif \mathbf{x}$.
Notice that after integrating over the spatial domain $\Omega$, the structure equation \eqref{eq:preserved_integrated} only depends on the temporal variable $t$.
Therefore, one introduces a time-dependent factor $R(t)$ by applying a variable transformation
\begin{equation*}
	u(\mathbf{x}, t) = R(t)  v(\mathbf{x}, t),
\end{equation*}
such that the structure equation \eqref{eq:preserved} can be merged into the PDE \eqref{eq:pde} to form an augmented PDE system as:
\begin{equation} \label{eq:tdsr}
		\begin{dcases}
			\dfrac{\partial}{\partial t} (R v) = \mathcal{A} [R v], \\
				\displaystyle \frac{\dif}{\dif t} \int_\Omega \mathcal{K}_{\mathcal{Q}}[R v](\mathbf{x},t) \dif \mathbf{x} = \int_\Omega \mathcal{K}_{\mathcal{S}}[R v](\mathbf{x},t) \dif \mathbf{x},
		\end{dcases}
\end{equation}
with $$R(0)=1,\quad v({\mathbf x},0) = u_0({\mathbf x}).$$
Since $R(t)$ can be treated as a constant within the spatial integral, we factor out $R(t)$ from the integrals of $\mathcal{K}_{\mathcal{Q}}$ and $\mathcal{K}_{\mathcal{S}}$ as
\begin{equation*}
	\begin{aligned}
		\int_\Omega \mathcal{K}_{\mathcal{Q}}[R  v](\mathbf{x}, t) \dif \mathbf{x} &= \mathcal{F}_{\mathcal{Q}}[R](t) \int_\Omega \mathcal{K}_{\mathcal{Q}}^v[v](\mathbf{x}, t) \dif \mathbf{x}, \\
		\int_\Omega \mathcal{K}_{\mathcal{S}}[R v](\mathbf{x}, t) \dif \mathbf{x} &= \mathcal{F}_{\mathcal{S}}[R](t) \int_\Omega \mathcal{K}_{\mathcal{S}}^v[v](\mathbf{x}, t) \dif \mathbf{x},
	\end{aligned}
\end{equation*}
where $\mathcal{F}_{\mathcal{Q}}, \, \mathcal{F}_{\mathcal{S}}: ([0, T] \to \mathbb{K}) \to ([0, T] \to \mathbb{K})$ are the factors depending on $R(t)$, and $\mathcal{K}_{\mathcal{Q}}^v, \, \mathcal{K}_{\mathcal{S}}^v: (\Omega \times [0, T] \to \mathbb{K}) \to (\Omega \times [0, T] \to \mathbb{K})$ are the renormalized integral kernels depending on $v(\mathbf{x}, t)$.
Therefore, the structure equation \eqref{eq:preserved_integrated} can be rewritten into an ordinary differential equation (ODE) for $R(t)$ as (referred to as structure ODE):
\begin{equation} \label{eq:tdsr_ode}
	\begin{dcases}
		\frac{\dif}{\dif t} \big( \mathcal{F}_{\mathcal{Q}}[R] \mathcal{I}_{\mathcal{Q}}[v] \big) = \mathcal{F}_{\mathcal{S}}[R]  \mathcal{I}_{\mathcal{S}}[v], \vspace{5pt}\\
		\mathcal{F}_{\mathcal{Q}}[R](0) \mathcal{I}_{\mathcal{Q}}[v](0) = C_0,
	\end{dcases}
 \end{equation}
	where
	\begin{equation*}
	\begin{aligned}
		\mathcal{I}_{\mathcal{Q}}[v](t) = \int_\Omega \mathcal{K}_{\mathcal{Q}}^v[v](\mathbf{x},t) \dif \mathbf{x}, \quad
		\mathcal{I}_{\mathcal{S}}[v](t) = \int_\Omega \mathcal{K}_{\mathcal{S}}^v[v](\mathbf{x},t) \dif \mathbf{x}.
	\end{aligned}
\end{equation*}
Here, $\mathcal{I}_{\mathcal{Q}}, \, \mathcal{I}_{\mathcal{S}}: (\Omega \times [0, T] \to \mathbb{K}) \to ([0, T] \to \mathbb{K})$ are the integral operators of the renormalized integral kernels $\mathcal{K}_{\mathcal{Q}}^v, \, \mathcal{K}_{\mathcal{S}}^v$.
Thus, by alternately solving the original PDE and the structure equation in \eqref{eq:tdsr}, the TDSR method helps the solutions adhere to intrinsic physical properties.
In the context of traditional numerical methods, the structure ODE \eqref{eq:tdsr_ode} is solved either by deriving the analytical solution \cite{cole2017tdsr} or by fixed-point iteration \cite{hou2024energy}.
This framework can deal with both conservative and dissipative systems, and allows a flexible cooperation of a wide range of PDE systems.

\subsection{Physics-informed neural networks}

PINN \cite{karniadakis2021nature, raissi2019physics} and its extensions \cite{Zhoutao2023Failure,  Zhaojia2021Solving,  pang2019fpinns, wang2024respecting, wang2021understanding} have become popular deep learning techniques for solving PDEs.
The vanilla PINN adopts a multi-layer perceptron (MLP) to approximate the solution function $u(\mathbf{x}, t)$, which is denoted as
\begin{equation} \label{eq:nn}
	\begin{split}
		\bar u_{\text{NN}}: \, & \mathbb{R}^{d+1} \to \mathbb{R}^n, \\
	& (\mathbf{x}, t) \mapsto \mathbf{W}_{\text{out}} \sigma(\mathbf{W}_L \sigma(\cdots \sigma(\mathbf{W}_1 (\mathbf{x}, t) + \mathbf{b}_1) \cdots) + \mathbf{b}_L) + \mathbf{b}_{\text{out}},
	\end{split}
\end{equation}
where the input vector is aligned as $(\mathbf{x}, t) = (x_1, x_2, \cdots, x_d, t)$.
The number of hidden layers is denoted as $L$, and the width (\ie, the number of neurons in each hidden layer) is denoted as $\{ W_i \}_{i = 1}^L$.
The weights and biases of the $i$-th hidden layer are denoted as $\mathbf{W}_{(i)} \in \mathbb{R}^{W_{i+1} \times W_i}$ and $\mathbf{b}_i \in \mathbb{R}^{W_{i+1}}$, respectively.
Additionally, the output layer is denoted as $\mathbf{W}_{\text{out}} \in \mathbb{R}^{n \times W_L}$ and $\mathbf{b}_{\text{out}} \in \mathbb{R}^n$ to match the output dimension.
Here $\sigma(\cdot)$ is the activation function. To ensure the smoothness of the solution, the activation function is often chosen as the hyperbolic tangent function $\sigma(\cdot) = \tanh(\cdot)$ in scientific computing.
During the implementation of PINN, the widths of the hidden layers are often chosen to be the same, \ie, $W_i = W$, $\forall\, i = 1, 2, \cdots, L$.

The loss function is often designed to minimize the mean square $L^2$-norm (also called MSE) of the PDE residual, \ie, the difference between the left-hand side and the right-hand side of the PDEs.
The initial and boundary conditions are also incorporated into the loss function to ensure that the solutions satisfy the given conditions.
The minimization process is called learning or training in NN methods.
For the parabolic PDE problem \eqref{eq:pde}, the loss function of PINNs can be written as
\begin{equation} \label{eq:pinn_loss}
		\begin{aligned}
		&\mathcal{L}_{\text{PINN}}[\bar u_{\text{NN}}] = \mathcal{L}_{\text{PDE}}[\bar u_{\text{NN}}] + \lambda_1\mathcal{L}_{\text{IC}}[\bar u_{\text{NN}}] + \lambda_2\mathcal{L}_{\text{BC}}[\bar u_{\text{NN}}], \\
		&\text{where} \quad
		\begin{dcases}
			\mathcal{L}_{\text{PDE}}[\bar u_{\text{NN}}]  = \frac{1}{N_{\text{PDE}}} \sum_{i=1}^{N_{\text{PDE}}} \left| \frac{\partial }{\partial t} \bar u_{\text{NN}}(\mathbf{x}_i, t_i) - \mathcal{A}[\bar u_{\text{NN}}](\mathbf{x}_i, t_i) \right|^2, \\
		\mathcal{L}_{\text{IC}}[\bar u_{\text{NN}}] = \frac{1}{N_{\text{IC}}} \sum_{j=1}^{N_{\text{IC}}} \left| \bar u_{\text{NN}}(\mathbf{x}_j, 0) - u_0(\mathbf{x}_j) \right|^2, \\
		\mathcal{L}_{\text{BC}}[\bar u_{\text{NN}}] = \frac{1}{N_{\text{BC}}} \sum_{k=1}^{N_{\text{BC}}} \left| \bar u_{\text{NN}}(\mathbf{x}_k, t_k) - g(\mathbf{x}_k, t_k) \right|^2.
		\end{dcases}
	\end{aligned}
\end{equation}
Here $\lambda_1>0$ and $\lambda_2>0$ are two balancing hyper-parameters, $\{ (\mathbf{x}_i, t_i) \}_{i=1}^{N_{\text{PDE}}} \in \Omega \times [0, T]$ are the collocation points for the PDE residual loss, and $ \{ \mathbf{x}_j\}_{j=1}^{N_{\text{IC}}} \in \Omega$ and $ \{ (\mathbf{x}_k, t_k) \}_{k=1}^{N_{\text{BC}}} \in \partial \Omega \times [0, T]$ are the collocation points for the initial and boundary conditions, respectively.
Notice that the loss function \eqref{eq:pinn_loss} depends solely on the given PDE formula and its associated conditions, without requiring any known solution data. This characteristic greatly enhances the practical applicability of PINN.

\begin{remark}
	NNs also can be designed to approximate operators or functionals \cite{chen1995universal}, enabling the development of operator-learning methods for PDEs \cite{kovachki2023neuraloperator, li2020FNO, lu2021deeponet}.
	These methods can either learn semi-discretized evolution operators \cite{li2020FNO}, or map from given conditions (such as initial conditions, boundary conditions, or coefficients) to solutions	\cite{lu2021deeponet}.
	Our proposed Sidecar framework can also be extended to these operator-learning methods to enhance physical fidelity.
	We leave the exploration of applying Sidecar to these methods as future work.
\end{remark}

There are some advanced techniques to improve the performance and training efficiency of PINN, such as the adaptive sampling strategy \cite{Zhoutao2023Failure, Zhaojia2021Solving} and the learning rate annealing algorithm \cite{wang2021understanding}. One insightful technique is the causal training strategy \cite{wang2024respecting}, which encourages PINN to learn the solution in accordance with the temporal causality of the PDEs.
To illustrate this idea, we discretize the time interval $[0, T]$ into $N_T>0$ sub-intervals with $\{ t^n \}_{n=0}^{N_T}$, and define the residual loss at each time step $t^n$ as
\begin{equation*}
	\mathcal{L}_{\text{PDE}}^n[\bar u_{\text{NN}}] = \frac{1}{N_{n}} \sum_{i=1}^{N_{n}} \left| \frac{\partial }{\partial t} \bar u_{\text{NN}}(\mathbf{x}_i, t^n) - \mathcal{A}[\bar u_{\text{NN}}](\mathbf{x}_i, t^n) \right|^2,
\end{equation*}
where $N_{n}$ is the number of spatial collocation points at time $t^n$. Therefore, the PDE residual loss can be written as $\mathcal{L}_{\text{PDE}}[\bar u] =  \frac{1}{N_T} \sum_{n=0}^{N_T} \mathcal{L}_{\text{PDE}}^n[\bar u]$. To further respect the temporal causality, the PDE residual loss is reformulated in a weighted approach:
\begin{equation} \label{eq:causal_loss}
	\tilde{\mathcal{L}}_{\text{PDE}}[\bar u_{\text{NN}}] =  \frac{1}{N_T} \sum_{n=0}^{N_T} w_n \, \mathcal{L}_{\text{PDE}}^n[\bar u_{\text{NN}}], \quad w_n = \exp \left( -\varepsilon \sum_{l = 0}^{n-1} \mathcal{L}_{\text{PDE}}^l[\bar u_{\text{NN}}] \right),
\end{equation}
where the temporal weight $w_n$ is set to be small unless all the previous time steps $\{ t^l \}_{0 \le l < n}$ have been well-approximated, and $\varepsilon$ is a hyper-parameter that controls the decay rate of the weights (\ie, the larger $\varepsilon$ indicates the higher accuracy requirement for the previous time steps). The causal training strategy can be easily integrated into the existing PINN solvers, and has shown great potential in improving the performance of PINN for parabolic PDEs with strong temporal dependencies.

\section{The Proposed Method}
\label{sec:methodology}

This section presents our proposed novel structure-preserving framework, Sidecar, to improve the physical consistency of the existing function-learning NN solvers, including network architecture, loss function, and training strategy.

\subsection{Network architecture}
\begin{figure}[t]
	\centering
	\includegraphics[width=0.8\textwidth]{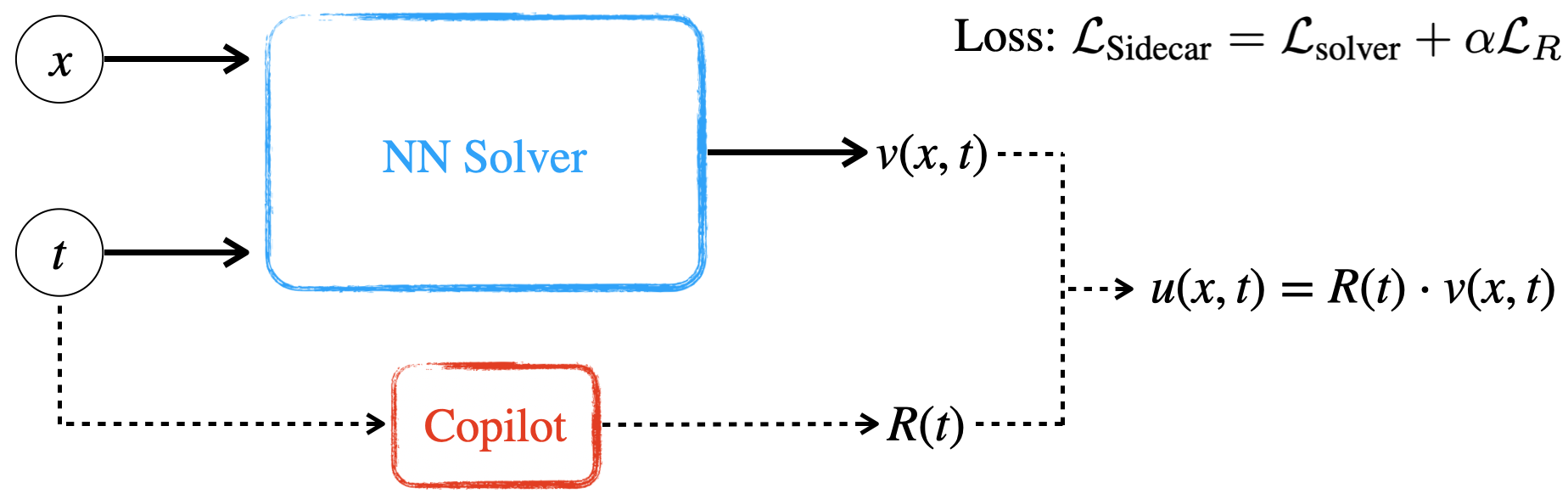}
	\vspace{-0.1cm}\caption{The network architecture of the proposed Sidecar framework.}
	\label{fig:sidecar}
\end{figure}

Inspired by the TDSR method,
we also rewrite the NN solution of the parabolic PDE problem \eqref{eq:pde} to be learned as
\begin{equation} \label{eq:sidecar}
	\bar u_{\text{NN}}(\mathbf{x}, t) = \bar R_{\text{NN}}(t) \, \bar v_{\text{NN}}(\mathbf{x}, t),
\end{equation}
where $\bar v_{\text{NN}}(\mathbf{x}, t)$ will be learned from the primary network, and $\bar R_{\text{NN}}(t)$ will be produced from the copilot network to guide the structure-preserving properties, and the overall solution $\bar u_{\text{NN}}(\mathbf{x}, t)$ satisfies the augmented  PDE system \eqref{eq:tdsr}.
The Sidecar framework is illustrated in \Cref{fig:sidecar}, where the \emph{primary-copilot} design allows it to be flexibly integrated with existing NN solvers.

As for the architecture choice of each part, the primary network $\bar v_{\text{NN}}(\mathbf{x}, t)$ can inherit the architecture of existing NN solvers, such as the MLP in \eqref{eq:nn} adopted in the vanilla PINN \cite{raissi2019physics}.
Meanwhile, to maintain computational efficiency and avoid overwhelming $\bar v_{\text{NN}}(\mathbf{x}, t)$, the copilot network $\bar R_{\text{NN}}(t)$ is implemented to be lightweight, such as a shallow MLP with significantly fewer neurons compared to $\bar v_{\text{NN}}(\mathbf{x}, t)$.

\subsection{Loss functions} \label{sec:loss}

Incorporating structure-preserving properties into network training requires a well-designed loss function, especially when the structure ODE \eqref{eq:tdsr_ode} contains spatial integral operators.
Here, we design the total loss to consist of two components:
\begin{equation} \label{eq:sidecar_loss}
	\mathcal{L}_{\text{Sidecar}}[\bar R_{\text{NN}}, \bar v_{\text{NN}}] = \mathcal{L}_{\text{solver}}[\bar R_{\text{NN}} \, \bar v_{\text{NN}}] + \alpha \mathcal{L}_{R}[\bar R_{\text{NN}}, \bar v_{\text{copy}}],
\end{equation}
where $\mathcal{L}_{\text{solver}}$ is the solver loss that guides the solution to learn the PDE solution, $\mathcal{L}_{R}$ is the structure loss that ensures the copilot network $\bar R_{\text{NN}}(t)$ effectively guides the primary network $\bar v_{\text{NN}}(\mathbf{x}, t)$ to adhere to the physical properties, and $\bar v_{\text{copy}}$ is a detached copy of $\bar v_{\text{NN}}$ (and will be discussed in this section later).
$\alpha>0$ is a hyper-parameter to control the relative weight, and is set to $\alpha = 1$ by default in our experiments.
Although fine-tuning $\alpha$ may lead to further improvements, the advantages of Sidecar remain robust to the choice of $\alpha$.
While $\mathcal{L}_{\text{Sidecar}}$ adopts the typical \emph{main-regularization} format, its implementation is specifically tailored for Sidecar.

\subsubsection{The solver loss}
We design $\mathcal{L}_{\text{solver}}[\bar R_{\text{NN}} \, \bar v_{\text{NN}}]$ to evaluate the combined output rather than primary network $\bar v_{\text{NN}}(\mathbf{x}, t)$ only, allowing the copilot network $\bar R_{\text{NN}}$ to learn PDE-related information.
The design of $\mathcal{L}_{\text{solver}}$ can be inherited from the chosen primary NN solver. In the case of vanilla PINN \cite{raissi2019physics}, $\mathcal{L}_{\text{solver}}$ is the PINN loss function as defined in Eq.~\eqref{eq:pinn_loss}:
\begin{equation} \label{eq:solver_loss}
	\mathcal{L}_{\text{solver}}[\bar R_{\text{NN}} \, \bar v_{\text{NN}}] = \mathcal{L}_{\text{PINN}}[\bar R_{\text{NN}} \, \bar v_{\text{NN}}].
\end{equation}
We also simply take the hyper-parameters $\lambda_1=\lambda_2=1$ in $\mathcal{L}_{\text{PINN}}$.

\subsubsection{The structure loss}
To bake in the structure-preserving properties, a direct but impractical approach is
to define the structure loss $\mathcal{L}_{R}$ as the residual of the structure ODE \eqref{eq:tdsr_ode}, where spatial integral operators $\mathcal{I}_{\mathcal{Q}}[\bar v_{\text{NN}}]$ and $\mathcal{I}_{\mathcal{S}}[\bar v_{\text{NN}}]$ are evaluated using numerical integration algorithms based on $\bar v_{\text{NN}}(\mathbf{x}, t)$.
However, this approach requires the algorithms to be differentiable for back-propagation.
Although several numerical integration algorithms exist, such as \texttt{torchquad} \cite{Gomez2021torchquad} or \texttt{torch.trapezoid} \cite{paszke2019pytorch}, they are mainly based on low-order numerical schemes and are invalid for complex PDE systems with discontinuities or singularities.
Meanwhile, the memory cost during back-propagation can become prohibitively large, particularly when dealing with a large number of collocation points.

To facilitate the incorporation of the existing high-accuracy numerical integration schemes, we propose to compute the integrals $\mathcal{I}_{\mathcal{Q}}$ and $\mathcal{I}_{\mathcal{S}}$
 based on $\bar v_{\text{copy}}(\mathbf{x}, t)$, a detached copy of $\bar v_{\text{NN}}(\mathbf{x}, t)$.
The adopted numerical integration algorithm is the Romberg integration \cite{gautschi2011numerical}, which is a widely used scheme with high accuracy.
Subsequently, we minimize the structure loss $\mathcal{L}_{R}$ with respect to $\bar R_{\text{NN}}(t)$ only, while keeping $\bar v_{\text{NN}}(\mathbf{x}, t)$ does not participate in the back-propagation of $\mathcal{L}_{R}$, \ie, $\min_{\bar R_{\text{NN}}} \mathcal{L}_{R}[\bar R_{\text{NN}}, \bar v_{\text{copy}}].$

To avoid back-propagating through the numerical integration, we need to temporally discretize the structure ODE \eqref{eq:tdsr_ode}.
For simplicity, we consider a regular grid for the time steps $t^n = n \cdot \delta t$, $n = 0, 1, \cdots, N_T$, where $\delta t = T / N_T$ and denote discrete variables and operators as
\begin{equation*}
	\begin{array}{lll}
		\bar R_{\text{NN}}^{n} := \bar R_{\text{NN}}(t^n), &
		\displaystyle
		\hat{\mathcal{I}}_{\mathcal{Q}} [v^{n}] = \int_\Omega \mathcal{K}_{\mathcal{Q}}^v[v^{n}](\mathbf{x}) \dif \mathbf{x}, \vspace{3pt}\\
		\bar v_{\text{copy}}^{n}(\mathbf{x}) := \bar v_{\text{NN}}(\mathbf{x}, t^n), &
		\displaystyle
		\hat{\mathcal{I}}_{\mathcal{S}} [v^{n}] = \int_\Omega \mathcal{K}_{\mathcal{S}}^v[v^{n}](\mathbf{x}) \dif \mathbf{x},
	\end{array}
\end{equation*}
where $\hat{\mathcal{I}}_{\mathcal{Q}}, \, \hat{\mathcal{I}}_{\mathcal{S}}: (\Omega \to \mathbb{K}) \to \mathbb{K}$ are the discrete version of the integral operators $\mathcal{I}_{\mathcal{Q}}, \, \mathcal{I}_{\mathcal{S}}$ in Eq.~\eqref{eq:tdsr_ode}, respectively.
Similarly, the discrete version of the factor $\mathcal{F}_{\mathcal{Q}}, \, \mathcal{F}_{\mathcal{S}}$ in \eqref{eq:tdsr_ode} are denoted as $\hat{\mathcal{F}}_{\mathcal{Q}}, \, \hat{\mathcal{F}}_{\mathcal{S}}: \mathbb{K} \to \mathbb{K}$, respectively.

\textbf{a)} For a conservative system (\ie, $\mathcal{S}[R  v] = 0$), the structure loss is designed as
\begin{equation} \label{eq:structure_loss_conservative}
	\mathcal{L}_{R}[\bar R_{\text{NN}}, \bar v_{\text{copy}}] = \frac{1}{N_T} \sum_{n=0}^{N_T} \left| \hat{\mathcal{F}}_{\mathcal{Q}}[\bar R_{\text{NN}}^n] \, \hat{\mathcal{I}}_{\mathcal{Q}}[\bar v_{\text{copy}}^n] - C_0 \right|^2,
\end{equation}
where the constant $C_0 = \mathcal{Q} \circ \iota[u_0]$ is given by the initial condition.

\textbf{b)} For dissipative systems (\ie, $\mathcal{S}[R  v] < 0$), we apply the backward Euler method to discretize the structure ODE \eqref{eq:tdsr_ode} as: for $n = 0, \cdots, N_T - 1$,
\begin{equation} \label{eq:backeuler}
\frac{ \hat{\mathcal{F}}_{\mathcal{Q}}[\bar R_{\text{NN}}^{n+1}] \, \hat{\mathcal{I}}_{\mathcal{Q}}[\bar v_{\text{copy}}^{n+1}] - \hat{\mathcal{F}}_{\mathcal{Q}}[\bar R_{\text{NN}}^{n}] \, \hat{\mathcal{I}}_{\mathcal{Q}}[\bar v_{\text{copy}}^{n}]}{\delta t} = \hat{\mathcal{F}}_{\mathcal{S}}[\bar R_{\text{NN}}^{n+1}] \, \hat{\mathcal{I}}_{\mathcal{S}}[\bar v_{\text{copy}}^{n+1}].
\end{equation}
Combined with the initial condition $\mathcal{F}_{\mathcal{Q}}[R](0) \, \mathcal{I}_{\mathcal{Q}}[v](0) = C_0$, we denote the residual of the discrete structure ODE \eqref{eq:backeuler} as:
\begin{equation*}
		\mathcal{L}_{R}^0[\bar R_{\text{NN}}, \bar v_{\text{copy}}] := \left(\hat{\mathcal{F}}_{\mathcal{Q}}[\bar R_{\text{NN}}^0] \, \hat{\mathcal{I}}_{\mathcal{Q}}[\bar v_{\text{copy}}^{0}] - C_0 \right)^2,
\end{equation*}
and for $n = 0, \cdots, N_T - 1$,
\begin{equation*}\small
		\mathcal{L}_{R}^{n+1}[\bar R_{\text{NN}}, \bar v_{\text{copy}}] := \left( \frac{\hat{\mathcal{F}}_{\mathcal{Q}}[\bar R_{\text{NN}}^{n+1}] \, \hat{\mathcal{I}}_{\mathcal{Q}}[\bar v_{\text{copy}}^{n+1}] - \hat{\mathcal{F}}_{\mathcal{Q}}[\bar R_{\text{NN}}^{n}] \, \hat{\mathcal{I}}_{\mathcal{Q}}[\bar v_{\text{copy}}^{n}]}{\delta t} - \hat{\mathcal{F}}_{\mathcal{S}}[\bar R_{\text{NN}}^{n+1}] \, \hat{\mathcal{I}}_{\mathcal{S}}[\bar v_{\text{copy}}^{n+1}] \right)^2.
\end{equation*}
Then the structure loss $\mathcal{L}_{R}$ can be defined using the residual $\mathcal{L}_{R}^n$ as:
\begin{equation} \label{eq:structure_loss_dissipative}
	\mathcal{L}_{R}[\bar R_{\text{NN}}, \bar v_{\text{copy}}] = \frac{1}{N_T} \sum_{n=0}^{N_T} \mathcal{L}_{R}^n[\bar R_{\text{NN}}, \bar v_{\text{copy}}].
\end{equation}
\begin{remark}
Inspired by the causal training strategy \cite{wang2024respecting}, we can reformulate the structure loss $\mathcal{L}_{R}$ as the weighted form to respect the temporal causality:
\begin{equation} \label{eq:causal_structure_loss}
	\begin{aligned}
	& \tilde{\mathcal{L}}_{R}[\bar R_{\text{NN}}, \bar v_{\text{copy}}] =  \frac{1}{N_T} \sum_{n=0}^{N_T} w_n \, \mathcal{L}_R^n[\bar R_{\text{NN}}, \bar v_{\text{copy}}],
	\end{aligned}
\end{equation}
with  $w_n = \exp \left( -\varepsilon \sum_{l = 0}^{n-1} \mathcal{L}_R^l[\bar R_{\text{NN}}, \bar v_{\text{copy}}] \right).$
\end{remark}

\begin{remark}
	To discretize the structure ODE \eqref{eq:tdsr_ode}, we adopt the backward Euler method for simplicity, while it can be easily extended to other time discretization schemes, such as Runge-Kutta methods and backward difference formula methods.
\end{remark}

\subsection{Training strategy}

Although the primary-copilot design of Sidecar facilitates flexible cooperation with existing NN solvers, it also introduces significant training challenges: Two separate networks must serve different purposes while maintaining consistency, all without incurring excessive computational cost.
We design the training procedure of Sidecar to ensure that the structure knowledge can enhance, rather than constrain, the learning process. It involves two stages:

\textbf{a)} \emph{Synchronization}: We equip the primary network $\bar v_{\text{NN}}(\mathbf{x}, t)$ with the copilot network $\bar R_{\text{NN}}(t)$, and train them  to minimize the solver loss $\mathcal{L}_{\text{solver}}$ in Eq.~\eqref{eq:solver_loss}, \ie,
\begin{equation*}
	\min_{\bar R_{\text{NN}}, \bar v_{\text{NN}}} \mathcal{L}_{\text{solver}}[\bar R_{\text{NN}} \, \bar v_{\text{NN}}].
\end{equation*}
During this stage, all training strategies inherited from the primary NN solver can be applied, such as the adaptive sampling \cite{Zhoutao2023Failure, Zhaojia2021Solving} and the causal training \cite{wang2024respecting}.
This stage allows the primary network $\bar v_{\text{NN}}(\mathbf{x}, t)$ to achieve sufficient accuracy, allowing a well-estimated spatial integral within the structure loss $\mathcal{L}_{R}$ in Eq.~\eqref{eq:structure_loss_conservative} or Eq.~\eqref{eq:structure_loss_dissipative}.
Additionally, this stage ensures that the copilot network $\bar R_{\text{NN}}(t)$ is synchronized with $\bar v_{\text{NN}}(\mathbf{x}, t)$, offering a reliable initialization for the next stage.

\textbf{b)} \emph{Navigation}: In addition to the solver loss $\mathcal{L}_{\text{solver}}$, the structure loss $\mathcal{L}_{R}$ is introduced and minimized with respect to $\bar R_{\text{NN}}(t)$ only, \ie,
\begin{equation*}
	\min_{\bar R_{\text{NN}}\bar v_{\text{NN}}}  \mathcal{L}_{\text{solver}}[\bar R_{\text{NN}} \, \bar v_{\text{NN}}] + \alpha \mathcal{L}_{R}[\bar R_{\text{NN}}, \bar v_{\text{copy}}] .
\end{equation*}
The second stage aims to navigate the learned solution $\bar R_{\text{NN}}(t) \bar v_{\text{NN}}(\mathbf{x}, t)$ to better satisfy the structure ODE \eqref{eq:tdsr_ode}, enhancing both accuracy and physical consistency.
As discussed in Section~\ref{sec:loss}, the spatial integrals $\mathcal{I}_{\mathcal{Q}}[\bar v_{\text{copy}}]$ and $\mathcal{I}_{\mathcal{S}}[\bar v_{\text{copy}}]$ within $\mathcal{L}_{R}$ are computed using a detached copy $\bar v_{\text{copy}}$, which is not involved in the back-propagation process.
This stage acts as a fine-tuning process
and thus require significantly fewer epochs compared to the first stage. If the first stage involves $K_1$ training epochs, the second stage typically uses $K_2 \ll K_1$ epochs.

The two-stage training is easy to implement by setting the coefficient $\alpha = 0$ in $\mathcal{L}_{\text{Sidecar}}[\bar R_{\text{NN}}, \bar v_{\text{NN}}] $ \eqref{eq:sidecar_loss} during the first stage and then updating it to $\alpha = 1$ for the second stage.
The overall training procedure for Sidecar is outlined in Algorithm \ref{alg:sidecar}.

\begin{algorithm}[t]
	\caption{The Training Strategy for the Proposed Sidecar Framework}
	\label{alg:sidecar}
	\begin{algorithmic}[0]
		\STATE \textbf{Input:} The PDE  ~\eqref{eq:pde} and its structure equation \eqref{eq:preserved}, the primary and copilot network architectures, the training epochs $K_1$ and $K_2$.
		\STATE \textbf{Output:} The  trained primary network $\bar v_{\text{NN}}(\mathbf{x}, t)$ and the copilot network $\bar R_{\text{NN}}(t)$.
		\STATE \# Stage 1: Synchronization
		\FOR {$k = 1$ to $K_1$}
			\STATE Train $\bar v_{\text{NN}}(\mathbf{x}, t)$ and $\bar R_{\text{NN}}(t)$ with the  loss $\mathcal{L}_{\text{solver}}[\bar R_{\text{NN}}\bar v_{\text{NN}}]$.
		\ENDFOR
		\STATE \# Stage 2: Navigation
		\FOR {$k = 1$ to $K_2$}
			\STATE Compute the integrals within $\mathcal{L}_{R}$ by a detached copy $\bar v_{\text{copy}}(\mathbf{x}, t)$ of $\bar v_{\text{NN}}(\mathbf{x}, t)$.
			\STATE Train $\bar v_{\text{NN}}(\mathbf{x}, t)$ and $\bar R_{\text{NN}}(t)$ with the loss $\mathcal{L}_{\text{solver}}[\bar R_{\text{NN}} \bar v_{\text{NN}}] + \mathcal{L}_{R}[\bar R_{\text{NN}}, \bar v_{\text{copy}}]$.
		\ENDFOR
	\end{algorithmic}
\end{algorithm}

\subsection{Applications to parabolic PDEs with structure-preserving properties} \label{sec:applications}

Here, we illustrate the application of the Sidecar framework to parabolic  PDEs with structure-preserving properties, including the NLS equation \eqref{eq:nls} and the AC equation \eqref{eq:allen_cahn}.

\subsubsection{Nonlinear Schrödinger equation}
To apply Sidecar to the NLS equation \eqref{eq:nls}, we first introduce the TDSR factor $R(t)$.
Since Eq.~\eqref{eq:nls} is a complex-valued PDE problem, the primary part $v(\mathbf{x}, t)$ is naturally a complex-valued function, but the TDSR factor $R(t)$ could be either $R(t) \in \mathbb{C}$ or $R(t) \in \mathbb{R}$.
Here, we choose a real-valued $R(t)$, and the overall solution can be written as
\begin{equation} \label{eq:nls_tdsr_real}
	u(\mathbf{x}, t) = R(t) \big( \re [v](\mathbf{x}, t) + i  \im [v](\mathbf{x}, t) \big).
\end{equation}
It enables rewriting the norm of $u$ as $|u|^2 = R^2 \cdot \left( \re [v]^2 + \im [v]^2 \right)$, and the real-valued form of the PDE \eqref{eq:nls} as
\begin{equation} \label{eq:nls_renormalized}
	\begin{dcases}
		\hfill - 2 R \cdot \im [v]_t - 2 R_t \cdot \im [v] + R \cdot \Delta \re [v] + 2 R^3 \left( \re [v]^2 + \im [v]^2 \right) \re [v] = 0, \\
	\hfill 2 R \cdot \re [v]_t + 2 R_t \cdot \re [v] + R \cdot \Delta \im [v] + 2 R^3 \left( \re [v]^2 + \im [v]^2 \right) \im [v] = 0,
	\end{dcases}
\end{equation}
with $R(0) = 1, v(\mathbf{x}, 0) = u_0(\mathbf{x})$.
After applying the transformation \eqref{eq:nls_tdsr_real}, the structure ODE of the mass conservation law \eqref{eq:nls_mass} and the momentum conservation law \eqref{eq:nls_momentum} can be rewritten as
\begin{subequations}
\begin{align}
	&R^2  \mathcal{I}_1[v] = C_0^{(1)}, \quad \mathcal{I}_1[v](t) = \int_\Omega |v(\mathbf{x},t)|^2 \dif \mathbf{x}, \label{eq:nls_mass_ODE} \\
	&R^2  \mathcal{I}_2[v] = C_0^{(2)}, \quad \mathcal{I}_2[v](t) = \int_\Omega \Big( \! \re[v] \im[\nabla v] - \im [v]  \re[\nabla v] \Big) \dif \mathbf{x}. \label{eq:nls_momentum_ODE}
\end{align}
\end{subequations}
Therefore, the structure loss can be written as
\begin{equation} \label{eq:nls_structure_loss}
	\mathcal{L}_{R}[\bar R_{\text{NN}}, \bar v_{\text{copy}}] = \frac{1}{N_T} \sum_{n=0}^{N_T} \left| \left( \bar R_{\text{NN}}^n \right)^2 \, \hat{\mathcal{I}}_{\mathcal{Q}_i}[\bar v_{\text{copy}}^n] - C_0^{(i)} \right|^2, \quad i = 1, 2,
\end{equation}
where $\hat{\mathcal{I}}_{\mathcal{Q}_i}$ are the discrete versions of the integral operators $\mathcal{I}_{\mathcal{Q}_i}$ in \eqref{eq:nls_mass_ODE} and \eqref{eq:nls_momentum_ODE}, respectively, and $C_0^{(i)}= \mathcal{Q}_i \circ \iota[u_0]$ for $i = 1, 2$.

\subsubsection{Allen--Cahn equation}
With $u(\mathbf{x}, t) = R(t)v(\mathbf{x}, t)$ the AC equation \eqref{eq:allen_cahn} can be rewritten as
\begin{equation} \label{eq:allen_cahn_renormalized}
	\begin{dcases}
		R \, v_t + R_t \, v = \varepsilon^2 R \, \Delta v + f[R v], \\
		\frac{\dif} {\dif t} \mathcal{E}_{AC}[R v] = \mathcal{S}_{AC}[R v],
	\end{dcases}
\end{equation}
with $R(0) = 1, v(\mathbf{x}, 0) = u_0(\mathbf{x})$, and $\mathcal{E}_{AC}[R v](0) = \mathcal{E}_{AC} \circ \iota[u_0]$.
Let us take $\mathcal{E}_{AC}$ and $\mathcal{S}_{AC}$ in \eqref{eq:allen_cahn_energy} with the double-well potential functional $F[u] = \frac{1}{4} (u^2 - 1)^2$, and $f = - F' = \frac{1}{2} (u - u^3)$. Then factoring out $R$ from integrals gives
\begin{align*}
	\mathcal{E}_{AC}[R  v] &= R^4 \int_\Omega \frac{1}{4} v^4 \dif \mathbf{x} + R^2 \int_\Omega \frac{1}{2} \left(\varepsilon^2 \nabla v^2 - v^2 \right) \dif \mathbf{x} + \int_\Omega \frac{1}{4} \dif \mathbf{x}, \\
	\mathcal{S}_{AC}[R v] &= - R_t^2 \int_\Omega v^2 \dif \mathbf{x} - R^2 \int_\Omega v_t^2 \dif \mathbf{x} - R_t R \int_\Omega 2 v v_t \dif \mathbf{x}.
\end{align*}
By omitting the constant term, the structure ODE can be derived as
\begin{equation} \label{eq:allen_cahn_renormalized_ode}
	\begin{dcases}
		\frac{\dif} {\dif t} \left( R^4 \, \mathcal{I}_{\mathcal{Q}, 1}[v] + R^2 \, \mathcal{I}_{\mathcal{Q}, 2}[v] \right) = R_t^2 \, \mathcal{I}_{\mathcal{S}, 1}[v] + R^2 \, \mathcal{I}_{\mathcal{S}, 2}[v] + R R_t \, \mathcal{I}_{\mathcal{S}, 3}[v], \\
		R^4(0) \, \mathcal{I}_{\mathcal{Q}, 1}[v](0) + R^2(0) \, \mathcal{I}_{\mathcal{Q}, 2}[v](0) = C_0,
	\end{dcases}
\end{equation}
where
\begin{align*}
	&\mathcal{I}_{\mathcal{Q}, 1}[v] = \int_\Omega \frac{1}{4} v^4 \dif \mathbf{x}, \quad
	\mathcal{I}_{\mathcal{Q}, 2}[v] = \int_\Omega \frac{1}{2} \left(\varepsilon^2 \nabla v^2 - v^2 \right) \dif \mathbf{x}, \\
	&\mathcal{I}_{\mathcal{S}, 1}[v] = - \int_\Omega v_{t}^2 \dif \mathbf{x}, \quad
	\mathcal{I}_{\mathcal{S}, 2}[v] = - \int_\Omega v^2 \dif \mathbf{x}, \quad
	\mathcal{I}_{\mathcal{S}, 3}[v] = - \int_\Omega 2 v v_t \dif \mathbf{x}.
\end{align*}
After applying the backward Euler method to discretize the structure ODE \eqref{eq:allen_cahn_renormalized_ode}, we obtain  the residual of the structure ODE as:
\begin{equation*}
	\mathcal{L}_{R}^0 [\bar R_{\text{NN}}, \bar v_{\text{copy}}] = \left( \left( \bar R_{\text{NN}}^0 \right)^4 \, \mathcal{I}_{\mathcal{Q}, 1}[\bar v_{\text{copy}}^0] + \left( \bar R_{\text{NN}}^0 \right)^2 \, \mathcal{I}_{\mathcal{Q}, 2}[\bar v_{\text{copy}}^0] - C_0 \right)^2,
\end{equation*}
and  for $n = 0, \ldots, N_T - 1$,
\begin{equation} \label{eq:allen_cahn_structure_loss}
	\begin{aligned}
	\mathcal{L}_{R}^{n+1} [\bar R_{\text{NN}}, \bar v_{\text{copy}}] &= \left( \bar R_{\text{NN}}^{n+1} \right)^4 \, \mathcal{I}_{\mathcal{Q}, 1}[v_{\text{copy}}^{n+1}] + \left( \bar R_{\text{NN}}^{n+1} \right)^2 \, \mathcal{I}_{\mathcal{Q}, 2}[\bar v_{\text{copy}}^{n+1}]\\
		& \quad \;- \left( \bar R_{\text{NN}}^{n} \right)^4 \, \mathcal{I}_{\mathcal{Q}, 1}[\bar v_{\text{copy}}^{n}]   - \left( \bar R_{\text{NN}}^{n} \right)^2 \, \mathcal{I}_{\mathcal{Q}, 2}[\bar v_{\text{copy}}^{n}] \\
	& \quad \; - \delta t \left( \! \big( \widetilde R_t^{n+1} \big)^2 \, \mathcal{I}_{\mathcal{S}, 1}[\bar v_{\text{copy}}^{n+1}] \right.\left.+ \left( \bar R_{\text{NN}}^{n+1} \right)^2 \mathcal{I}_{\mathcal{S}, 2}[\bar v_{\text{copy}}^{n+1}] \right.\\
	&\qquad\qquad\left.+ \widetilde R_{t}^{n+1} \bar R_{\text{NN}}^{n+1} \mathcal{I}_{\mathcal{S}, 3}[\bar v_{\text{copy}}^{n+1}] \right)^2,
	\end{aligned}
\end{equation}
where $\widetilde R_{t}^{n+1} = (\bar R_{\text{NN}}^{n+1} - \bar R_{\text{NN}}^{n}) / \delta t$ is the difference quotient of $R(t)$. The discrete structure ODE \eqref{eq:allen_cahn_renormalized_ode} for the AC equation \eqref{eq:allen_cahn} has a more complicated form, mainly due to the dissipation speed $\mathcal{S}_{AC}[R  v]$ in \eqref{eq:allen_cahn_dissipation} involving the temporal derivative of $R(t)$. Finally  the structure loss $\mathcal{L}_{R}[\bar R_{\text{NN}}, \bar v_{\text{copy}}]$ can be defined  through \eqref{eq:structure_loss_dissipative}.

\section{Numerical Experiments}
\label{sec:experiments}
In this section, we present some experiments on the NLS equation \eqref{eq:nls} and the AC equation \eqref{eq:allen_cahn} to demonstrate the effectiveness of the Sidecar framework in enhancing NN solvers with the structure-preserving capability and accuracy.

\subsection{Experimental setup}

For each case, we select a typical primary NN solver and implement both its vanilla and Sidecar-enhanced versions. For the Sidecar-enhanced version, the primary network $\bar v_{\text{NN}}(x, t)$ and the copilot network $\bar R_{\text{NN}}(t)$ are parameterized by two NNs, respectively. For the MLP case, the primary NN solver uses width $W_v$ and depth $L_v$, while the copilot network uses width $W_R$ and depth $L_R$, with $W_R \ll W_v$ and $L_R \ll L_v$ since the copilot network is expected to be small and lightweight.
For a fair comparison, we also ensure that the total number of neurons in the vanilla and compared Sidecar-enhanced versions is the same;
thus each vanilla version is configured with depth $\widehat L_v = L_v$ and width $\widehat W_v = W_v + W_R L_R / L_v$ so that both vanilla  and
Sidecar-enhanced versions have a similar number of neurons in total. We refer to this as an equivalent vanilla version. 

Both vanilla and Sidecar-enhanced versions are trained with the same training data and hyper-parameters for the loss functions.
The training data $ \{ \mathbf{x}_j \}_{j=1}^{N_{\text{IC}}}$
 and $ \{ (\mathbf{x}_k, t_k) \}_{k=1}^{N_{\text{BC}}}$
are equally spaced collocation points for the initial and boundary conditions, respectively, while the PDE residual points $ \{ (\mathbf{x}_i, t_i) \}_{i=1}^{N_{\text{PDE}}} \in \Omega \times [0, T]$ are the collocation points in the inner domain $\Omega$.
The test set used to evaluate the performance of the trained models is $2 \times$refined from the training set.
The number of training epochs $K_0$ of the equivalent vanilla version is set to be equal to the sum of that of the two stages of the Sidecar-enhanced version in  Algorithm \ref{alg:sidecar}, \ie, $K_0 = K_1 + K_2$. The detailed hyper-parameters of Sidecar are summarized in \Cref{tab:hyper-parameters} for all test problems.

Their performance is evaluated using the solver loss, the $L^2$ error with respect to the reference solution (\ie, $\left\| \bar u_{\text{NN}} - u \right\|_2$), and the $L^\infty$ error of the structure-preserving property as $\max_{t \in [0,T]} \left| \mathcal{Q}[\bar u_{\text{NN}}](t) - \mathcal{Q}[u](t) \right|$.
The code is implemented with the \texttt{PyTorch} library \cite{paszke2019pytorch}, while it can be easily extended to other deep learning frameworks such as \texttt{JAX} \cite{jax2018github}.
The experiments are conducted on an NVIDIA A100 GPU.
Each experiment is repeated 10 times with different random seeds, and the results are averaged over these runs. The shaded areas in the error plots represent the trust intervals with a confidence level of 95\%.

\begin{table}[thbp!] \label{tab:hyper-parameters}\small
	\begin{center}
	\caption{The hyper-parameters of the Sidecar-enhanced networks for all tested problems.}
	  \begin{tabular}{c|c|c|c}
		& \textbf{1D NLS} & \textbf{1D AC} & \textbf{2D AC} \\
				\hline
		$N_{\text{IC}}$ & 512 & 512 & 50 \\
		$N_{\text{BC}}$ & 128 & 200 & 50 \\
		$N_{\text{PDE}}$ & 65,536 & 10240 & 128$\times$128$\times$24 \\
		\hline
		$L_v$ & 4 & 4 & 6 \\
		$W_v$ & 50 \, 100 \, 200 \, 400 & 64 \, 128 \, 256 & 128 \\
		\hline
		$L_R$ & 2 & 2 & 2\\
		$W_R$ & 10 & 16 & 16\\
		\hline
		$K_1$ & 100,000 & 180,000 & 4,000 per interval \\
		$K_2$ & 20,000 & 20,000 & 1,000 per interval \\
	  \end{tabular}
	\end{center}
  \end{table}

\subsection{Numerical results}

Here we report the numerical results of applying Sidecar to the NLS and AC equations.

\begin{figure}[t!]
	\centering
	\includegraphics[width=\textwidth]{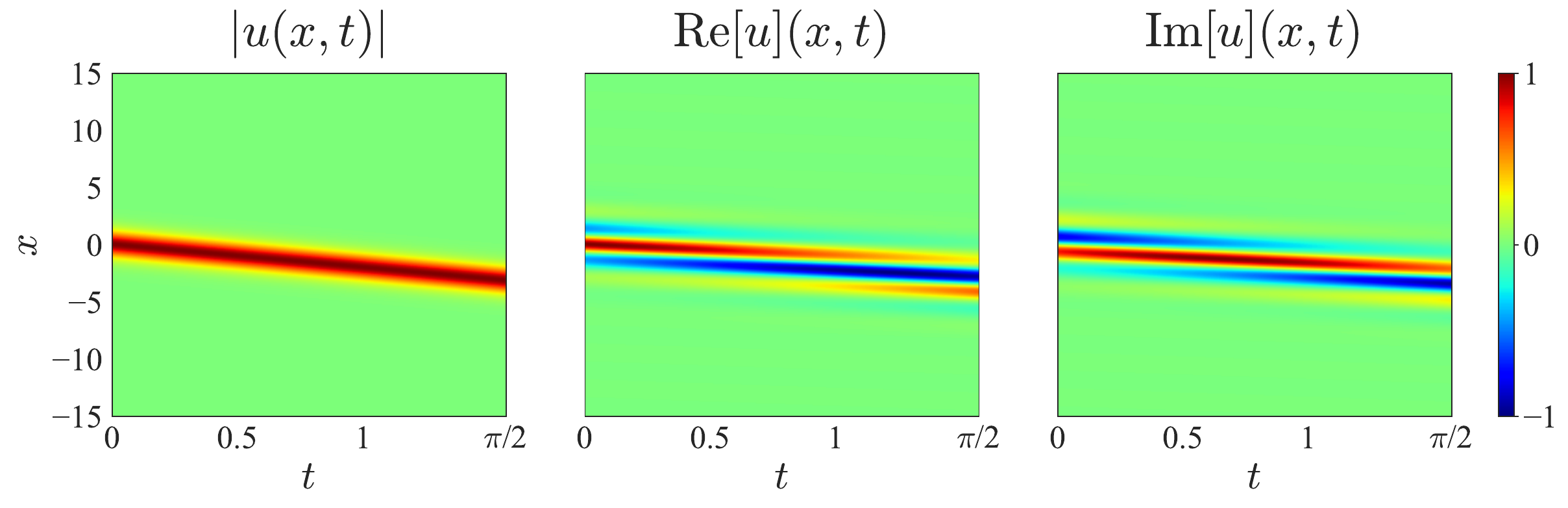}
	\vspace{-0.7cm}\caption{\Cref{ex:nls_mass} -- Illustration of the moving soliton solution for the 1D NLS equation \eqref{eq:nls_1D}.}
	\label{fig:nls_solution}
\end{figure}

\begin{example}[1D NLS Equation] \label{ex:nls_mass}
Consider the following 1D NLS equation
\begin{equation} \label{eq:nls_1D}
	\begin{dcases}
		-i u_t = \frac{1}{2} u_{xx} + |u|^2 u, & (x, t) \in [-15, 15] \times (0, \pi/2], \\
		u(x, 0) = u_0(x), & x \in [-15, 15], \\
		u(-15, t)  = u(15, t), \: u_x(-15, t) = u_x(15, t), & t \in [0, \pi/2].
	\end{dcases}
\end{equation}
We assume the moving soliton solution, a typical solution to the NLS equation describing a stable and localized wave packet that propagates without changing shape \cite{debnath2005nonlinear}. The initial condition and the corresponding exact solution are then given by
\begin{equation} \label{eq:nls_solution}
	u_0(x) = \sech (x) e^{-2ix}, \quad u(x, t) = \sech (x + 2t) \, e^{ -i (2x + \frac{3}{2} t )}.
\end{equation}
See \Cref{fig:nls_solution} for an illustration of the moving soliton solution of the 1D NLS equation \eqref{eq:nls_1D}.
\end{example}

\begin{figure}[t]
	\centering
	\includegraphics[width=\textwidth]{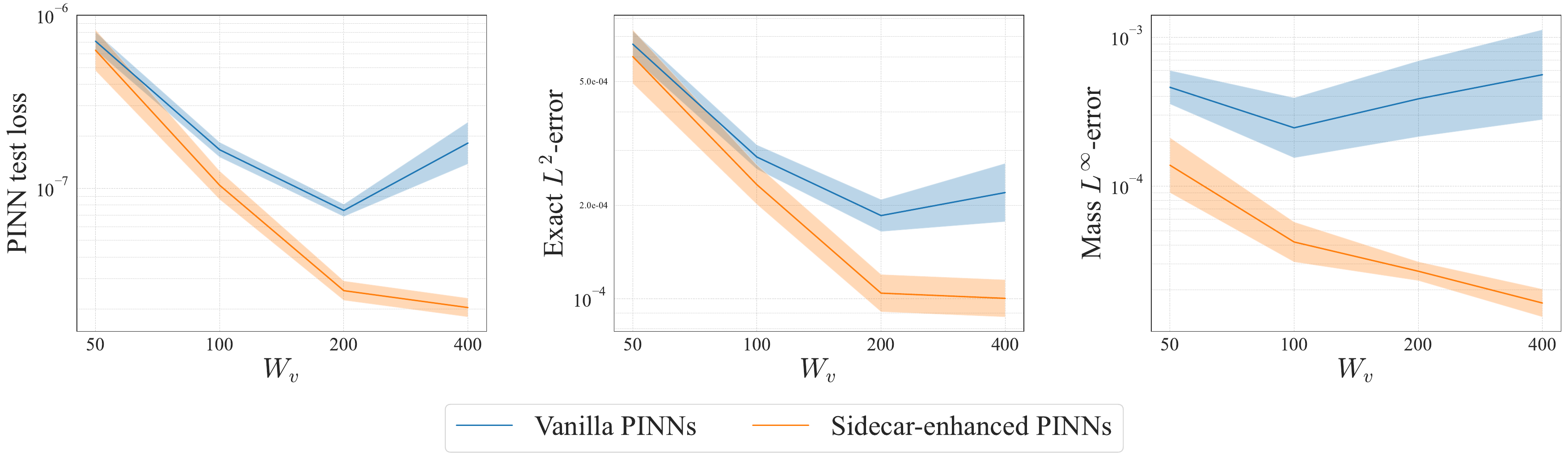}
	\vspace{-0.5cm}\caption{\Cref{ex:nls_mass} -- Performance comparisons between the Sidecar-enhanced PINNs with mass conservation and the equivalent vanilla PINNs for the 1D NLS equation \eqref{eq:nls_1D}. From left to right: the PINN test loss, the $L^2$ error of the numerical solution, and the $L^\infty$ error of the mass conservation.}
	\label{fig:nls1}
\end{figure}

\paragraph{Sidecar with mass conservation}
We first consider the mass conservation law \eqref{eq:nls_mass_ODE}, where the total mass $C_0^{(1)} = 2 \tanh(15)$ is obtained due to the initial condition \eqref{eq:nls_solution}.
We take into account mass conservation and build the corresponding  Sidecar-enhanced PINN for solving the 1D NLS equation \eqref{eq:nls_1D}.
Algorithm  \ref{alg:sidecar} with the PINN loss $\mathcal{L}_{\text{PINN}}$ \eqref{eq:pinn_loss} is used for the learning process.
The performance is compared with the equivalent vanilla PINN and the results are shown in \Cref{fig:nls1} in terms of the PINN test loss, the $L^2$ error of the numerical solution, and the $L^\infty$ error of the mass conservation. The Sidecar-enhanced PINNs result in more accurate solutions compared to the equivalent vanilla PINNs, while also better preserving the mass conservation property.
As illustrated in the right panel of \Cref{fig:nls1}, the numerical mass in the equivalent vanilla PINN does not converge with increasing network width, whereas the Sidecar-enhanced PINN leverages the larger network width to achieve better solution accuracy and mass conservation.

\begin{figure}[t!]
	\centering
	\includegraphics[width=\textwidth]{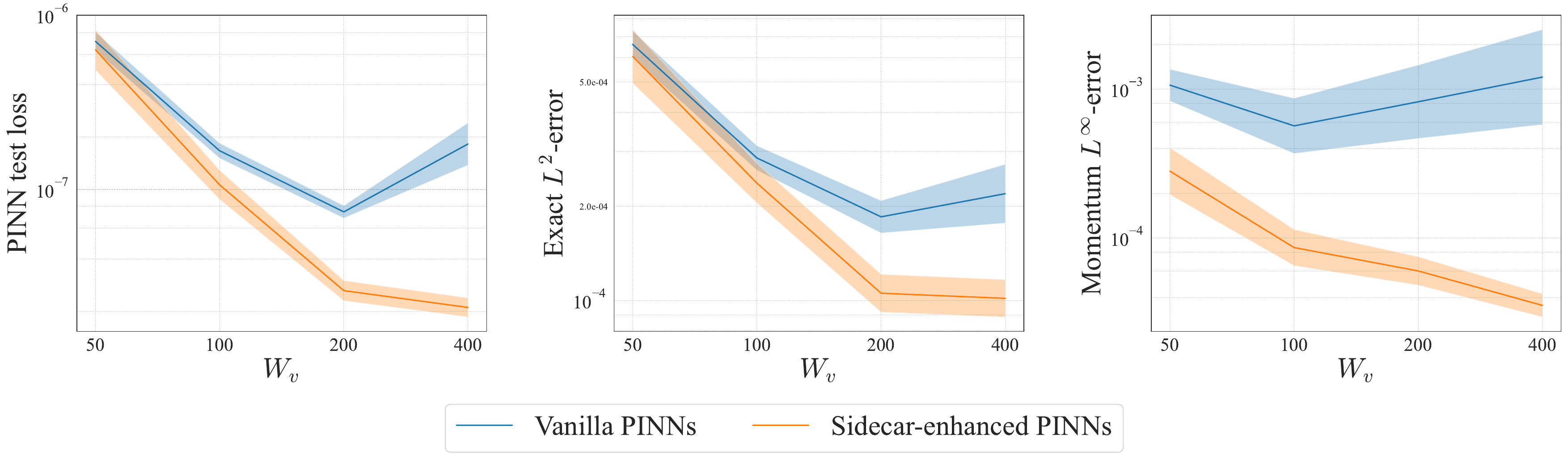}
	\vspace{-0.5cm}\caption{\Cref{ex:nls_mass} -- Performance comparisons between the Sidecar-enhanced PINNs with momentum conservation and the equivalent vanilla PINNs for the 1D NLS equation \eqref{eq:nls_1D}. From left to right: the PINN test loss, the $L^2$ error of the numerical solution, and the $L^\infty$ error of the momentum conservation.}
	\label{fig:nls2}
\end{figure}

\begin{figure}[t]
	\centering
	\includegraphics[width=0.7\textwidth]{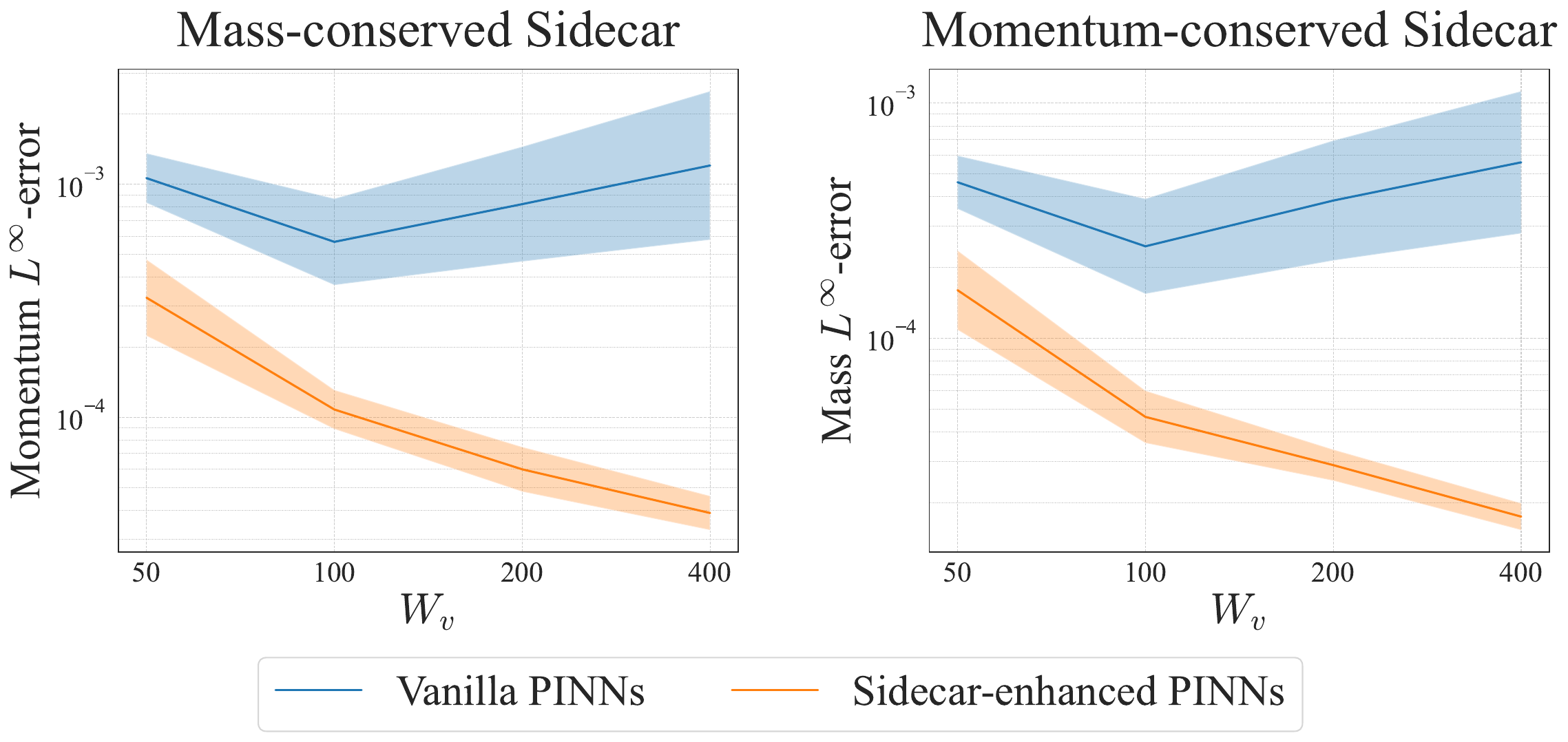}
	\vspace{-0.1cm}\caption{
	 \Cref{ex:nls_mass} -- Left: comparisons on the momentum conservation performance between the Sidecar-enhanced PINNs with mass conservation and the equivalent vanilla PINNs; right: comparisons on the mass conservation performance between the Sidecar-enhanced PINNs with momentum conservation and the equivalent vanilla PINNs for the NLS equation \eqref{eq:nls_1D}.}
	\label{fig:nls3}
\end{figure}

\paragraph{Sidecar with momentum conservation}
We next consider the momentum conservation law \eqref{eq:nls_momentum_ODE} and incorporate it into the Sidecar framework (instead of the mass conservation), where the total momentum $C_0^{(2)} = -4\tanh(15)$ is based on the initial condition \eqref{eq:nls_solution}.
 We compare the performance of the Sidecar-enhanced PINN with an equivalent vanilla PINN and
the results are shown in \Cref{fig:nls2} in terms of the PINN test loss, the $L^2$ error of the numerical solution, and the $L^\infty$ error of the momentum conservation, which indicate that the Sidecar enhances both solution accuracy and momentum conservation.
Moreover, since the NLS equation has both mass and momentum conservation laws, we further investigate the momentum conservation performance of the Sidecar-enhanced PINNs trained with the mass conservation law only and vice versa. The results are shown in \Cref{fig:nls3}, and
we observe that the Sidecar-enhanced PINNs have no trade-off between the mass and momentum conservation performance, i.e., either one of them used in the Sidecar-enhanced PINNs can help improve both of them and solution accuracy.

\begin{example}[1D AC Equation] \label{ex:ac_1D}
We apply Sidecar to solve the 1D AC equation considered in \cite{wang2024respecting}, which is given by
\begin{equation} \label{eq:allen_cahn_1D}
	\begin{dcases}
		u_t = \varepsilon^2 \, u_{xx} + 5(u - u^3), & (x, t) \in [-1, 1] \times [0, 1], \\
		u(x, 0) = x^2 \cos(\pi x), & x \in [-1, 1], \\
		u(-1, t) = u(1, t), \quad u_x(-1, t) = u_x(1, t), & t \in [0, 1],
	\end{dcases}
\end{equation}
where $\varepsilon = 0.01$.
Since an analytic solution is not available, we compute a high-resolution reference solution with a finite-difference method, which is plotted in \Cref{fig:allen_cahn_solution}(left).

\begin{figure}[t]
	\centering
	\includegraphics[width=\textwidth]{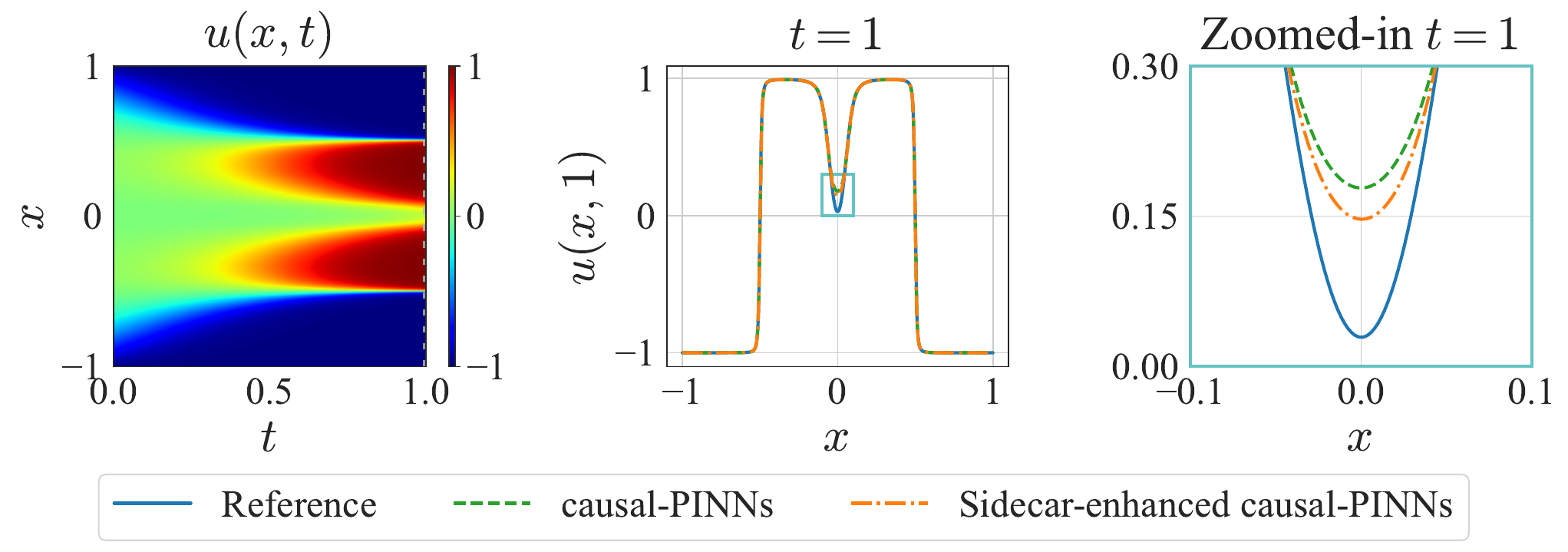}
	\vspace{-0.5cm}\caption{\Cref{ex:ac_1D} -- The solutions of the 1D AC equation \eqref{eq:allen_cahn_1D}.
	   {Left:} Illustration of the reference solution.
	{Middle:} Comparison of the snapshot at $t = 1$ from the reference solution, the Sidecar-enhanced causal-PINNs, and the equivalent causal-PINNs solutions \cite{wang2024respecting}.
	The cyan box indicates the region where the solution is zoomed in on the right.
	{Right:} The zoomed-in view of the solutions snapshot. The shown results are the worst cases of the 10 runs.}
	\label{fig:allen_cahn_solution}
\end{figure}

\paragraph{Sidecar with energy dissipation}
We choose the primary NN solver as proposed in \cite{wang2024respecting}, which involves the causal training strategy to reformulate the PDE loss into $\tilde{\mathcal{L}}_{\text{PDE}}$ \eqref{eq:causal_loss} for Algorithm  \ref{alg:sidecar}.
The Fourier feature embedding \cite{dong2021method} is not included in the primary network.
We refer to this solver as the causal-PINN, which is a variant of the vanilla PINN.
For the copilot network of Sidecar, we still use a lightweight MLP.
The structure loss $\mathcal{L}_{R}$ is derived from the structure ODE residual (i.e., energy dissipation) \eqref{eq:allen_cahn_structure_loss}.
The results are shown in \Cref{fig:allen_cahn} in terms of the causal-PINN test loss, the $L^2$ error of the numerical solution, and the $L^\infty$ error of the energy dissipation. Compared to the causal-PINNs, the Sidecar-enhanced causal-PINNs achieve better solution accuracy and energy dissipation.
This demonstrates that the proposed Sidecar framework can also be integrated with other primary NN solvers, showcasing its flexibility.
\begin{figure}[t!]
	\centering
	\includegraphics[width=\textwidth]{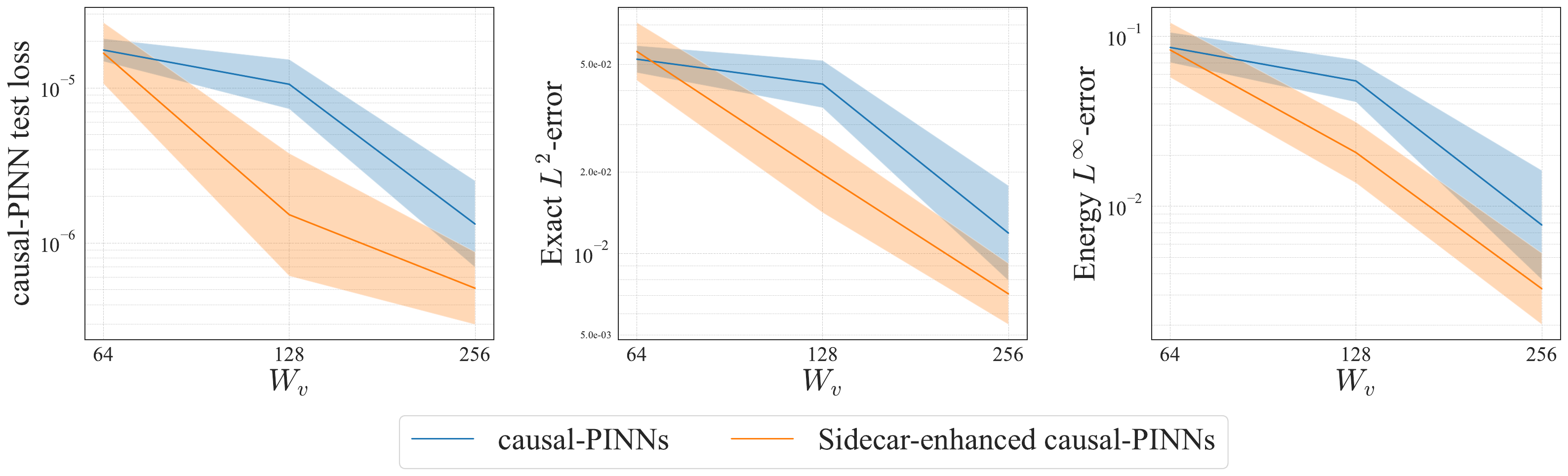}
	\vspace{-0.5cm}\caption{\Cref{ex:ac_1D} -- Comparisons of the Sidecar-enhanced causal-PINNs and the equivalent causal-PINNs \cite{wang2024respecting} for the 1D AC equation \eqref{eq:allen_cahn_1D}. From left to right: the causal-PINN test loss, the $L^2$ error of the numerical solution, and the $L^\infty$ error of the energy dissipation.}
	\label{fig:allen_cahn}
\end{figure}
\end{example}

\begin{example}[2D AC Equation] \label{ex:ac_2D}
We consider a benchmark problem for the 2D AC equation, known as the shrinking bubble problem \cite{chen1998applications} with the form
\begin{equation} \label{eq:allen_cahn_2D}
	\begin{dcases}
	u_t = \mu (\varepsilon^2 \Delta u + u - u^3), & (\mathbf{x}, t) \in \Omega \times [0, T], \\
	\frac{\partial u}{\partial \mathbf{n}} = 0, & (\mathbf{x}, t) \in \partial \Omega \times [0, T],
	\end{dcases}
\end{equation}
where $\mathbf{n}$ is the outward normal vector on the boundary $\partial \Omega$.
Here we choose the 2D domain $\Omega = [0, 1]^2$, the terminal time $T=10$, $\varepsilon = 0.025$, and $\mu = 10$ by following the settings in \cite{kutuk2024energy}.
The initial condition is given as
\begin{equation*}
u(\mathbf{x}, 0) = \tanh \left( \frac{0.35 - \sqrt{(x_1 - 0.5)^2 + (x_2 - 0.5)^2}}{2 \varepsilon} \right).
\end{equation*}
As a phase-field model, we focus on the evolution of the interface $\Gamma[t]:= \{ \mathbf{x} \in \Omega: u(\mathbf{x}, t) = 0 \}$, which is a closed curve in
the sharp interface limit,
which can be parameterized by the normalized arc-length parameter $s \in [0, 1]$.
The initial interface $\Gamma[0]$ forms a bubble-like shape centered at $\mathbf{x}_0 = (0.5, 0.5)$ with radius $r_0 = 0.35$.
Driven by interfacial tension, the bubble shrinks over time and eventually disappears.
During the evolution, $\Gamma[t]$ is expected to maintain radial symmetry, and its radius $r(t)$ is approximately described by
\begin{equation} \label{eq:allen_cahn_radius}
	r(t) = \sqrt{r_0^2 - 2 \varepsilon^2 t}.
\end{equation}
This approximation is valid when $\varepsilon$ is sufficiently small (i.e., $r(t) \gg \varepsilon$), as discussed in \cite{hou2024linear, ju2015fast}. The evolution of the shrinking bubble solution is visualized in \Cref{fig:allen_cahn_2D_solution}.

We extract the interface $\bar \Gamma[t]$ of a numerical solution $\bar u_{\text{NN}}(\cdot, t) \in \Omega$ using the marching squares algorithm \cite{lorensen1998marching}.
At each point $\mathbf{x} = \bar \Gamma[t](s)$ on the interface, we compute the Euclidean distance from the bubble center $\mathbf{x}_0$ as $\bar r(s, t) = \|\mathbf{x} - \mathbf{x}_0\|_2$,
and define the average $\bar r(t)$ and standard deviation $SD(t)$ as
\begin{equation*}
	\bar r(t) = \int_{\Gamma[t]} \bar r(s, t) \, \dif s, \quad SD(t) = \sqrt{ \int_{\Gamma[t]} \big( \bar r(s, t) - \bar r(t) \big)^2 \, \dif s}.
\end{equation*}
Here, the average radius $\bar r(t)$ is compared to the prediction $r(t)$ from \eqref{eq:allen_cahn_radius}, while a smaller $SD(t)$ indicates that the bubble remains close to a perfect circle. We use these metrics to evaluate the performance of the NN solvers in capturing the interface evolution and the energy dissipation property.

\begin{figure}[t]
	\centering
	\includegraphics[width=\textwidth]{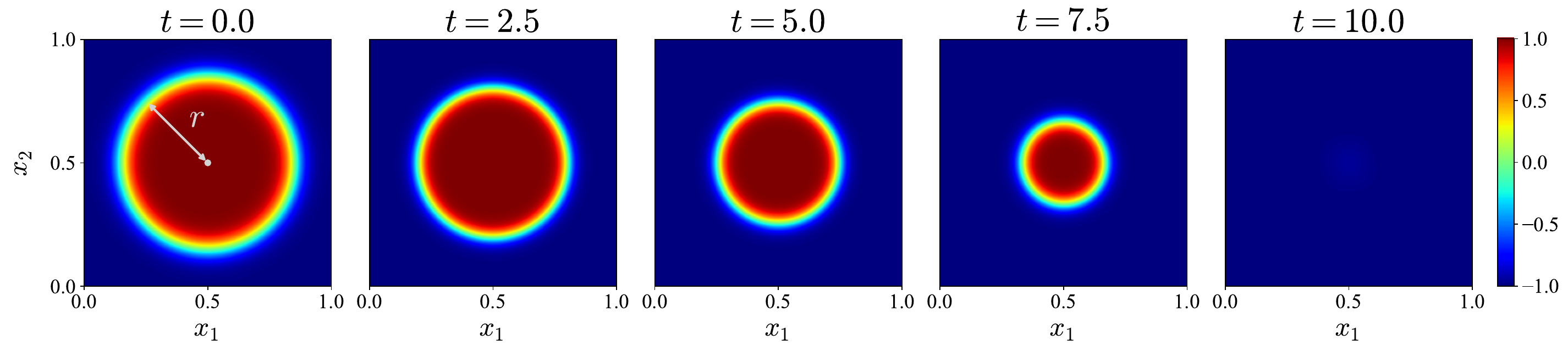}
	\vspace{-0.5cm}\caption{\Cref{ex:ac_2D} -- Illustration of the shrinking bubble solution of the 2D AC equation \eqref{eq:allen_cahn_2D}.}
	\label{fig:allen_cahn_2D_solution}
\end{figure}

\paragraph{Sidecar with energy dissipation}
We adopt the primary NN solver from \cite{kutuk2024energy}, which adopts a Residual Network (ResNet) architecture \cite{he2016deep}. We refer to this solver as ResNet-PINN.
To enforce the energy dissipation property, ResNet-PINN incorporates a differentiable regularization term into the loss function.
Adaptive approaches of the training point sampling and the temporal domain \cite{Zhaojia2021Solving} are involved to enhance performance. In each time interval, a separate ResNet is trained with the parameters initialized from the previous interval, following the transfer learning strategy \cite{zhuang2020comprehensive}.
We use ResNet-PINN as the primary network $\bar v_{\text{NN}}(\mathbf{x}, t)$ in Sidecar, incorporating all the techniques in \cite{kutuk2024energy} except for the energy dissipation regularization, which is already been considered by our Sidecar framework.
The copilot network $\bar R_{\text{NN}}(t)$ is implemented with a lightweight ResNet. The loss function is designed according to Sidecar for vanilla PINN with the structure loss $\mathcal{L}_{R}$ derived from the energy dissipation law \eqref{eq:allen_cahn_structure_loss} of the AC equation.
The training procedure adopts the transfer learning strategy from \cite{kutuk2024energy}, which divides the time interval $[0, T]$ into several sub-intervals, where each sub-interval is trained with a separate ResNet using the two-stage Sidecar approach. All hyper-parameters are listed in \Cref{tab:hyper-parameters} as stated.
The results are shown in \Cref{fig:allen_cahn_2D} in terms of the PDE-residual
loss, the standard deviation SD(t) away from a perfect circle, and the simulated radius.
The Sidecar-enhanced ResNet-PINNs achieve lower PDE residual loss and smaller standard deviation $SD(t)$, while well maintaining the evolution of radius $\bar r(t)$. This indicates that Sidecar can further enhance the performance of ResNet-PINNs by incorporating the energy dissipation law.
As the time $t \to 10$, both methods show some deviation from the theoretical prediction, which is expected since \eqref{eq:allen_cahn_radius} becomes invalid when $r(t)$ is comparable to $\varepsilon$, as discussed in \cite{hou2024linear, ju2015fast}.
\begin{figure}[t!]
	\centering
	\includegraphics[width=\textwidth]{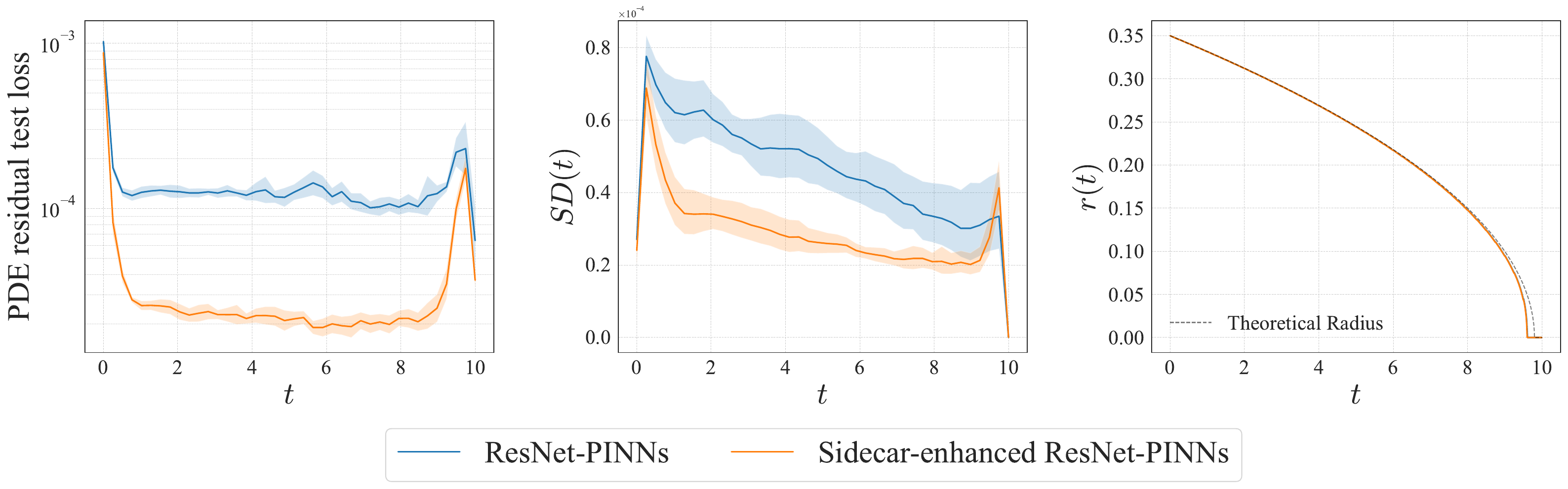}
	\vspace{-0.5cm}\caption{\Cref{ex:ac_2D} -- Comparisons of the Sidecar-enhanced ResNet-PINNs and the equivalent ResNet-PINNs \cite{kutuk2024energy} for the 2D AC equation in \Cref{ex:ac_2D}. From left to right: the PDE-residual loss, the standard deviation $SD(t)$ away from a perfect circle, and the simulated radius.}
	\label{fig:allen_cahn_2D}
\end{figure}
\end{example}

\section{Further Discussions}
\label{sec:discussions}

This section further discusses the reason why the proposed Sidecar framework can enhance the performance of existing NN solvers.
Ablation studies are conducted to validate the main components of Sidecar, including the architecture design and the loss design.
Specifically, we are interested in whether Sidecar benefits from:
\begin{enumerate}
	\item improving the representation capacity of NNs via the Sidecar architecture, and
	\item incorporating the structure-preserving knowledge via the loss design.
\end{enumerate}
We conduct a series of ablation studies to validate the above hypotheses.

\subsection{Representation capacity of the Sidecar architecture}
\label{sec:representation_capacity}

In Sidecar, the copilot network $\bar R_{\text{NN}}(t)$ is designed to be $t$-dependent only.
This architectural choice enables the network to more effectively model temporal dynamics in PDE solutions, potentially enhancing the overall representational capacity for temporal PDE solutions compared to standard MLPs used in vanilla PINNs.
To test this hypothesis, we train the primary network with and without a copilot network to directly fit the reference solution of a given PDE system, \ie,
\begin{equation*}
	\min_{\bar u_{\text{NN}}} \, \mathcal{L}_{\text{approx}}[\bar u_{\text{NN}}] = \frac{1}{N_{\text{PDE}}} \sum_{i=0}^{N_{\text{PDE}}} \big| \bar u_{\text{NN}}(\mathbf{x}_i, t_i) - u(\mathbf{x}_i, t_i) \big|^2,
\end{equation*}
where $\bar u_{\text{NN}}(\mathbf{x}, t)$ is the overall network output, which can be either a vanilla MLP or a copilot-equipped MLP.
By focusing solely on the function approximation task and removing the influence of the PDE solver loss, we can isolate and assess the impact of the Sidecar architecture on the network's ability to represent temporal structures.

\paragraph{Experimental settings}
We equip an MLP with a smaller MLP as a copilot, both of which are trained to approximate the reference solution of the 1D NLS equation \eqref{eq:nls_solution} and the 1D AC equation \eqref{eq:allen_cahn_1D}.
The performance is compared to an equivalent vanilla MLP in terms of the $L^2$ distance to the reference solution.
For a fair comparison, both versions are trained with the same number of randomly sampled training points (32,000 for NLS and 41,000 for AC) and epochs (10,000 for both cases).

\begin{figure}[t]
	\centering
	\includegraphics[width=0.7\textwidth]{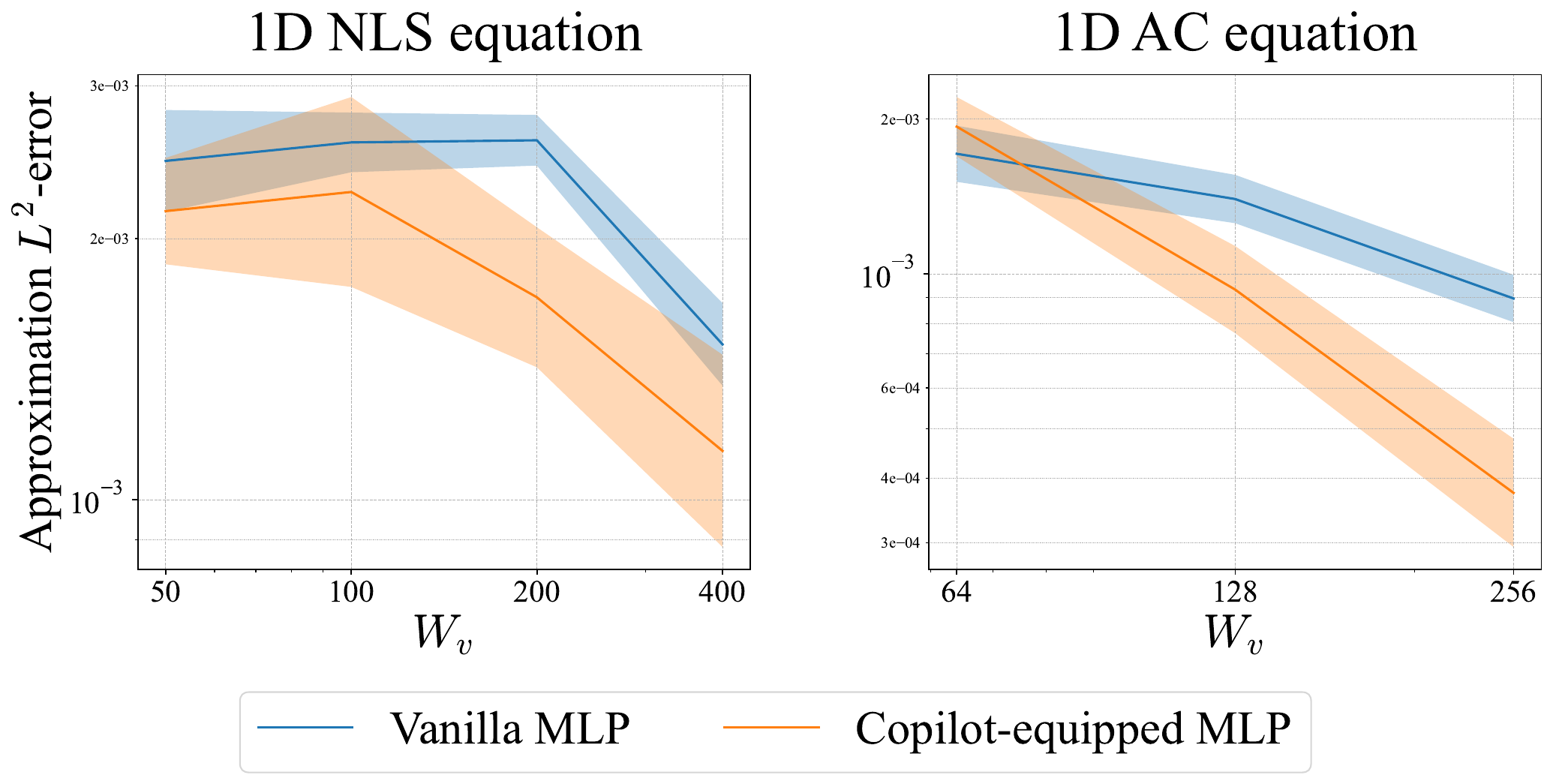}
	\vspace{-0.2cm}\caption{
		Comparison of copilot-equipped MLPs and the equivalent vanilla MLPs for approximating the reference solutions of the 1D NLS equation \eqref{eq:nls_1D} in \Cref{ex:nls_mass} and the 1D AC equation \eqref{eq:allen_cahn_1D} in \Cref{ex:ac_1D}.}
	\label{fig:representation_capacity}
\end{figure}

The results of $L^2$ errors of numerical solutions are shown in \Cref{fig:representation_capacity}, where the copilot-equipped primary networks consistently outperform their equivalent vanilla counterparts. This supports the hypothesis that the Sidecar architecture improves the representation capacity for parabolic PDE solutions.

With the same number of neurons and layers, the copilot-equipped primary network uses fewer parameters than the equivalent primary network but achieves higher accuracy. This shows that architecture tailored to the problem structures could outperform standard designs.
Similarly, Sidecar can be viewed as a PDE-friendly architecture, tailored for PDE systems with temporal evolution structures.
This idea aligns with recent works that introduce novel architectures specifically designed for NN solvers, such as the modified MLP \cite{wang2021understanding} and the volume weighted PINN \cite{song2025vw}.
However, these approaches do not explicitly incorporate physical structure information, which is a key feature of the Sidecar framework.

\subsection{Necessity of the structure loss}

Here, we validate the necessity of the structure loss $\mathcal{L}_{R}$ in incorporating structure-preserving knowledge.
Ideally, $\mathcal{L}_{R}$ should complement the PDE-based solver loss $\mathcal{L}_{\text{solver}}$ by explicitly embedding structure-preserving properties into the training process.
However, since these properties are inherently consistent with the PDE formula, it is possible that improvements in structure-preserving performance could be achieved using $\mathcal{L}_{\text{solver}}$ alone.

\paragraph{Experimental settings} We compare a Sidecar-enhanced PINN with and without the structure loss $\mathcal{L}_{R}$ during the second training stage, as well as the equivalent vanilla PINN for \Cref{ex:nls_mass}. All settings follow \Cref{tab:hyper-parameters}.
\begin{figure}[t]
	\centering
	\includegraphics[width=\textwidth]{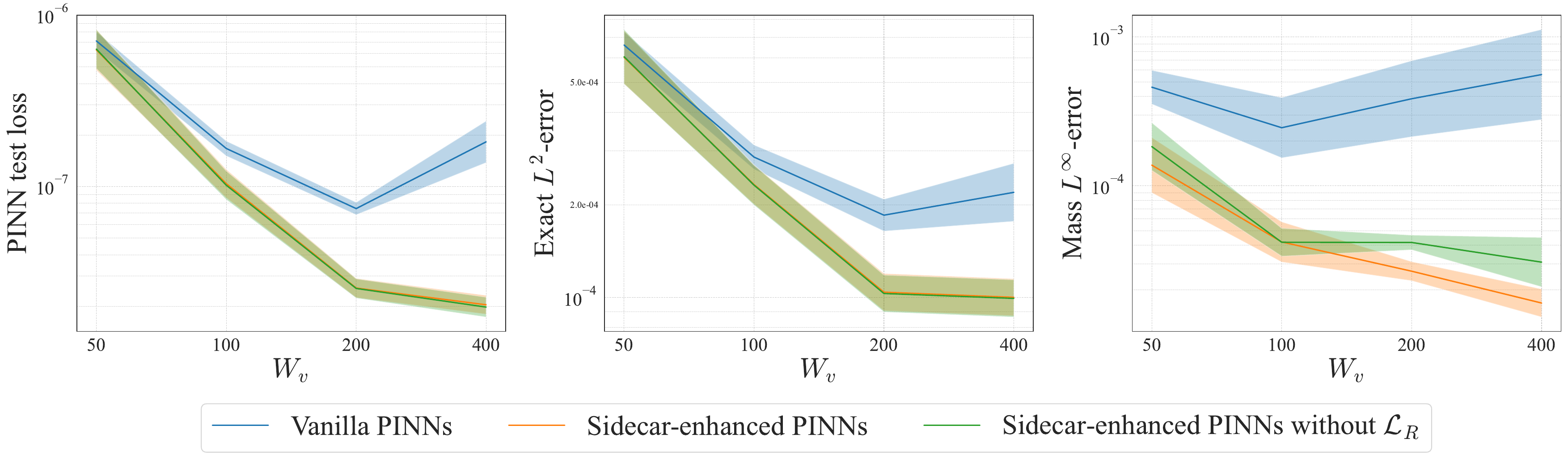}
	\vspace{-0.4cm}\caption{
	Comparison of the Sidecar-enhanced PINNs with and without the structure loss $\mathcal{L}_{R}$, and the equivalent vanilla PINNs \cite{raissi2019physics} for 1D NLS equation \eqref{eq:nls_1D} in \Cref{ex:nls_mass}. From left to right: the PINN test loss, the $L^2$ error of the numerical solution, and the $L^\infty$ error of the mass conservation.}
	\label{fig:necessity_structure_loss}
\end{figure}

The results are shown in \Cref{fig:necessity_structure_loss} in terms of the PINN test loss, the $L^2$ error of the numerical solution, and the $L^\infty$ error of the mass conservation. We can see that the Sidecar architecture and loss work together to improve the solution accuracy and physical fidelity.
The Sidecar-enhanced PINNs without the structure loss $\mathcal{L}_{R}$ outperform the vanilla PINNs, agreeing with the representational capacity improvement discussed in Section~\ref{sec:representation_capacity}.
Additionally, adding the $\mathcal{L}_{R}$  loss further improves the mass conservation law, particularly for larger network widths.
This highlights the critical role of $\mathcal{L}_{R}$.
Notably, adding $\mathcal{L}_{R}$ in Sidecar does not compromise the solution accuracy.
This distinguishes Sidecar from existing regularization-based NN solvers \cite{huang2024mass, kutuk2024energy}, which often encounter a trade-off between accuracy and physical fidelity.
In the Sidecar framework, this trade-off is prevented by the two-stage training and loss design, making it successful for integrating structure knowledge into NN solvers.

\section{Conclusion}
\label{sec:conclusion}
In this paper, we introduce a structure-preserving framework, Sidecar,  designed to cooperate with existing NN solvers for solving parabolic PDEs.
The framework combines a primary network with a lightweight copilot network, trained jointly to minimize a PDE-based solver loss $\mathcal{L}_{\text{solver}}$ and a structure loss $\mathcal{L}_{R}$.
The structure loss explicitly incorporates the system's structure-preserving properties, ensuring solutions adhere to intrinsic physical laws.
A two-stage training procedure is employed to first synchronize the two networks and then navigate the learned solution to respect structure-preserving properties.
Our Sidecar is flexible, compatible with existing NN solvers, and applicable to a wide range of parabolic PDE systems with different physical properties.
Experiments on the NLS and AC equations numerically demonstrate the effectiveness of Sidecar in improving solution accuracy and physical fidelity.
Ablation studies further validate the effectiveness of Sidecar's key components, showing improvements in representation capacity and the necessity of the structure loss $\mathcal{L}_{R}$ for embedding structure-preserving properties.

Regarding future work, we will extend Sidecar to more complex time-dependent PDE systems and NN solvers.
Advanced techniques for NN solvers, such as the strict embedding of boundary conditions \cite{dong2021method}, can be integrated into Sidecar to enhance its performance and applicability.
We will explore the simultaneous enforcement of multiple structure-preserving properties within the Sidecar framework to further enhance physical fidelity.
The preservation of local structural properties \cite{mu2018efficient} will also be investigated through some random tricks.
Application to operator-learning NN solvers, such as the Fourier Neural Operator \cite{li2020FNO}, is another promising direction.

\section*{Data availability}

The code and data used in this study are available at \url{https://github.com/DanclaChen/Sidecar}.

\section*{Acknowledgments}
The authors would like to extend their gratitude to Prof. Zuowei Shen and Prof. Qianxiao Li of National University of Singapore, as well as Prof. Yongqiang Cai of Beijing Normal University, for their helpful discussions and suggestions.

\bibliographystyle{siamplain}
\bibliography{refs}
\end{document}

%% file: ex_shared.tex
\usepackage{lipsum}
\usepackage{amsfonts}
\usepackage{graphicx}
\usepackage{epstopdf}
\usepackage{algorithmic}
\ifpdf
  \DeclareGraphicsExtensions{.eps,.pdf,.png,.jpg}
\else
  \DeclareGraphicsExtensions{.eps}
\fi

\graphicspath{{figures/}}

\newsiamremark{remark}{Remark}
\newsiamremark{example}{Example}
\newsiamremark{hypothesis}{Hypothesis}
\crefname{hypothesis}{Hypothesis}{Hypotheses}
\newsiamthm{claim}{Claim}
\newsiamremark{fact}{Fact}
\crefname{fact}{Fact}{Facts}

\usepackage{mathtools}

\newcommand*{\dif}{\mathop{}\!\mathrm{d}}
\newcommand*{\ie}{\emph{i.e.}}

\DeclareMathOperator{\re}{\operatorname{Re}}
\DeclareMathOperator{\im}{\operatorname{Im}}

\DeclareMathOperator{\sech}{\operatorname{sech}}

\headers{A Structure-Preserving Framework for Parabolic PDEs}{G. Chen, L. Ju, and Z. Qiao}

\title{A Structure-Preserving Framework for Solving Parabolic Partial Differential Equations with Neural Networks\thanks{Submitted to the editors DATE.
\funding{L. Ju is supported by the US Department of Energy grant DE-SC0025527. Z. Qiao's work is partially supported by the Hong Kong Research Grants Council RFS grant RFS2021-5S03, GRF grant 15305624 and NSFC/RGC Joint Research Scheme grant N\_PolyU5145/24.}}}

\author{Gaohang Chen\thanks{Department of Applied Mathematics, The Hong Kong Polytechnic University, Hung Hom, Kowloon, Hong Kong
  (\email{gaohang.chen@connect.polyu.hk}).}
\and Lili Ju\thanks{Department of Mathematics, University of South Carolina, Columbia, SC 29208, USA
  (\email{ju@math.sc.edu}).}
\and Zhonghua Qiao\thanks{Department of Applied Mathematics, The Hong Kong Polytechnic University, Hung Hom, Kowloon, Hong Kong
  (\email{zqiao@polyu.edu.hk}).}}

\usepackage{amsopn}
